\documentclass[
    textheight=9in,
    textwidth=5.6in,
    top=1in,
    headheight=12pt,
    headsep=25pt,
    footskip=30pt
  ]{article}

\usepackage[left=1in, right=1in, bottom=1in, top=1in]{geometry}

\usepackage[utf8]{inputenc} %
\usepackage[T1]{fontenc}    %
\usepackage{hyperref}       %
\usepackage{url}            %
\usepackage{booktabs}       %
\usepackage{amsfonts}       %
\usepackage{nicefrac}       %
\usepackage{microtype}      %
\usepackage{graphicx}
\usepackage[round]{natbib}  %

\usepackage{amsmath,amsfonts,bm}
\usepackage{mathtools}

\newcommand{\ba}[1]{\begin{align}#1\end{align}}
\newcommand{\bastar}[1]{\begin{align*}#1\end{align*}}

\newcommand{\given}{\,|\,}

\def\eqref#1{equation~\ref{#1}}

\def\1{\bm{1}}

\def\vzero{{\bm{0}}}

\def\vmu{{\bm{\mu}}}
\def\valpha{{\bm{\alpha}}}
\def\vtheta{{\bm{\theta}}}
\def\vxi{{\bm{\xi}}}

\def\vs{{\bm{s}}}

\def\vx{{\bm{x}}}
\def\vy{{\bm{y}}}
\def\vz{{\bm{z}}}

\def\mI{{\bm{I}}}

\DeclareMathAlphabet{\mathsfit}{\encodingdefault}{\sfdefault}{m}{sl}
\SetMathAlphabet{\mathsfit}{bold}{\encodingdefault}{\sfdefault}{bx}{n}

\def\gL{{\mathcal{L}}}

\def\gN{{\mathcal{N}}}

\newcommand{\E}{\mathbb{E}}

\DeclareMathOperator*{\argmin}{arg\,min}

\DeclarePairedDelimiterX{\infdivx}[2]{(}{)}{%
  #1\;\delimsize\|\;#2%
}
\newcommand{\kldiv}{D_{\mathrm{KL}}\infdivx}

 \newcommand{\epsilonv}[0]{\ensuremath{\boldsymbol{\epsilon}} }

\usepackage{amsmath}
\usepackage{amssymb}
\usepackage{mathtools}
\usepackage{amsthm}
\usepackage[capitalize,noabbrev]{cleveref}
\usepackage{multirow}
\usepackage{enumitem}
\usepackage{algorithm}
\usepackage{algorithmic}
\renewcommand{\algorithmicrequire}{\textbf{Input:}}
\renewcommand{\algorithmicensure}{\textbf{Output:}}
\usepackage[dvipsnames]{xcolor}
\definecolor{burntorange}{RGB}{191, 87, 0}

\usepackage{changepage}
\usepackage{capt-of}
\usepackage{caption}

\usepackage{setspace}  %
\usepackage{parskip}   %
\title{\textbf{Diffusion Boosted Trees}}

\author{ {\hspace{1mm}Xizewen Han and Mingyuan Zhou} \\
 \texttt{xizewen.han@utexas.edu,~~mingyuan.zhou@mccombs.utexas.edu} \\
	The University of Texas at Austin \\
	Austin, TX 78712 \\
}

\date{}

\hypersetup{
pdftitle={Diffusion Boosted Trees},
pdfsubject={stat.ML, cs.LG, stat.CO, stat.ME},
pdfauthor={Xizewen Han, Mingyuan Zhou},
pdfkeywords={diffusion models, boosting algorithm},
}

\begin{document}

\setlength{\abovedisplayskip}{3pt}
\setlength{\belowdisplayskip}{3pt}

\maketitle

\begin{abstract}
\noindent Combining the merits of both denoising diffusion probabilistic models and gradient boosting, the diffusion boosting paradigm is introduced for tackling supervised learning problems.
We develop Diffusion Boosted Trees (DBT), which can be viewed as both a new denoising diffusion generative model parameterized by decision trees (one single tree for each diffusion timestep), and a new boosting algorithm that combines the weak learners into a strong learner of conditional distributions without making explicit parametric assumptions on their density forms. We demonstrate through experiments the advantages of DBT over deep neural network-based diffusion models as well as the competence of DBT on real-world regression tasks, and present a business application (fraud detection) of DBT for classification on tabular data with the ability of learning to defer.
\end{abstract}

\section{Introduction}
\label{sec:intro}
A series of pivotal works in recent years \citep{song2019scorematching, ho2020ddpm, song2021scoresde, dhariwal2021diffbeatsgan, rombach2021stablediff, karras2022edm} has propelled diffusion-based generative models \citep{jsd2015diffusion} to the forefront of generative AI, capturing a significant amount of academic and industrial interest by the success of this class of models in content generation. Meanwhile, another line of work, Classification and Regression Diffusion Models (CARD) \citep{han2022card}, has been proposed to tackle supervised learning problems with a denoising diffusion probabilistic modeling framework, shedding new lights on both the foundational machine learning paradigm and the new elite in the generative AI family.

More specifically, CARD learns the target conditional distribution of the response variable $\vy$ given the covariates $\vx$, $p(\vy\given\vx)$, without imposing explicit parametric assumptions on %
its probability density function, and makes predictions by utilizing the stochastic nature of its output to directly generate samples that resemble $\vy$ from this target distribution. This framework has demonstrated outstanding results on both regression and image classification tasks: in regression, it shows the capability of modeling conditional distributions with flexible statistical attributes, and achieves state-of-the-art metrics on real-world datasets; for image classification, it introduces a novel paradigm to evaluate instance-level prediction confidence besides improving the prediction accuracy by a deterministic classifier.

However, CARD models are parameterized by deep neural networks. %
The work of
\citet{grinsztajn2022outperform} has %
illustrated that tree-based models remain the state-of-the-art function choice for modeling tabular data, and could outperform neural networks by a wide margin. \textit{Tabular data} is a crucial type of dataset for many supervised learning tasks, characterized by its table-format structure similar to a spreadsheet or a relational database, where each row represents an individual record or observation, and each column represents a feature or attribute of that record. Importantly, the features of tabular datasets are heterogeneous, including various types such as numerical (discrete or continuous) and categorical (nominal or ordinal), enabling the representation of diverse information about each record. This contrasts with image data, where the raw information is solely represented as pixel values.
CARD has not addressed classification tasks on tabular data, which represents an essential class of supervised learning tasks with wide applications in many areas. Therefore, significant potential remains to enhance the CARD framework to establish it as a universally applicable method in the realm of supervised~learning.

In this work, we aim to improve the CARD framework by incorporating trees as its function choice: trees are another vital class of universal approximators besides neural networks \citep{watt2020mlr, nisan1994boolfunc, hornik1989ffnnunivapprox}, and offer several advantages, including the automatic handling of missing values without the need for imputation, no requirement for data normalization during preprocessing, effective performance with less data, better interpretability, and robustness to outliers and irrelevant features. Additionally, we fill an important gap by applying the framework to classification on tabular data, which was not explored in the experiments presented by \citet{han2022card}. We start this quest by studying one of the most powerful supervised learning paradigms parameterized by trees: gradient boosting \citep{friedman2001gbm}.

Our main contributions are summarized as follows:
\begin{itemize}[noitemsep, topsep=0pt, leftmargin=*]
    \item We establish the connections between diffusion-based generative models and gradient boosting, a classic ensemble method for function estimation.
    \item We develop the \textbf{Diffusion Boosting} paradigm for supervised learning, which is simultaneously 1) a new denoising diffusion generative model that can be parameterized by decision trees --- a single tree for each diffusion timestep --- with a novel sequential training paradigm; and 2) a new boosting algorithm that combines the weak learners into a strong learner of conditional distributions without any assumptions on their parametric forms.
    \item Through experiments, we demonstrate that \textbf{D}iffusion \textbf{B}oosting \textbf{T}rees (DBT), the tree-based parameterization of our proposed paradigm, outperforms CARD on piecewise-defined functions and datasets with a large number of categorical features, while achieving competitive results in real-world regression tasks. DBT also excels in several other key areas: it offers interpretability at each diffusion timestep, maintains robust performance in the presence of missing data, and acts as an effective binary classifier on tabular data, featuring the ability to defer decisions with adjustable confidence levels.

\end{itemize}

\section{Background}
\label{sec:background}
We contextualize gradient boosting as a method for tackling \textit{supervised learning} tasks. Given a set of covariates $\vx=\{x_1, \dots, x_p\}$ and a response variable $\vy$, we seek to learn a mapping $F$ that takes $\vx$ as input and predicts $\vy$ as its output. It is common practice to impose a parametric form $\vtheta$ on $F$, casting supervised learning as a \textit{parameter optimization} problem: \ba{\vtheta^*=\argmin_{\vtheta} \Phi(\vtheta)=\argmin_{\vtheta} \E_{p(\vx, \vy)}\big[L\big(\vy, F(\vx; \vtheta)\big)\big], \label{eq:param_opt}} where $\vtheta^*$ is achieved by minimizing the expected value of some loss function $L(\vy, F)$. When gradient descent \citep{cauchy1847} is used to find the descent direction during the numerical optimization procedure, the optimal parameter is: \ba{\vtheta^*=\vtheta_0 + \sum_{m=1}^M\rho_m\cdot\big(-\nabla_{\vtheta_{m-1}}\Phi(\vtheta_{m-1})\big), \label{eq:optim_param_grad_desc}} where $M$ is the total number of update steps, $\vtheta_0$ is the initialization, and $\rho_m$ is the step size.

\subsection{Gradient Boosting}
\label{ssec:gradient_boosting}
While gradient descent can be described as a numerical optimization method \textit{in the parameter space}, gradient boosting \citep{friedman2001gbm} is essentially gradient descent \textit{in the function space}. With the objective function at the instance level, \ba{\Phi\big(F(\vx)\big)=\E_{p(\vy\given\vx)}\big[L\big(\vy, F(\vx)\big)\big],} by considering $F(\vx)$ evaluated at each $\vx$ as a parameter, its gradient can be computed as \ba{\nabla_{F(\vx)}\Phi\big(F(\vx)\big)=\frac{\partial \Phi\big(F(\vx)\big)}{\partial F(\vx)} =\E_{p(\vy\given\vx)}\left[\frac{\partial L\big(\vy, F(\vx)\big)}{\partial F(\vx)}\right], \label{eq:instance_function_grad}}
assuming sufficient regularity to interchange differentiation and integration. Following the gradient-based numerical optimization paradigm as in Eq.\,(\ref{eq:optim_param_grad_desc}), we obtain the optimal solution in the function space: \ba{F^*(\vx)=f_0(\vx) + \sum_{m=1}^M\rho_m\cdot\big(-g_m(\vx)\big), \label{eq:optim_func_grad_desc}} where $f_0(\vx)$ is the initial guess, and $g_m(\vx)=\nabla_{F_{m-1}(\vx)}\Phi\big(F_{m-1}(\vx)\big)$ is the gradient at optimization step $m$.

Given a finite set of samples $\{\vy_i, \vx_i\}_1^N$ from $p(\vx, \vy)$, we have the data-based analogue of $g_m(\vx)$ defined only at these training instances: $g_m(\vx_i)=\frac{\partial L\big(\vy_i, \hat{F}_{m-1}(\vx_i)\big)}{\partial \hat{F}_{m-1}(\vx_i)}.$ Since the goal of supervised learning is to generalize the predictive function $F$ to unseen data, \citet{friedman2001gbm} proposes to use a parameterized class of functions $h(\vx; \valpha)$ to estimate the negative gradient term for any $\vx$ at every gradient descent step. Specifically, $h(\vx; \valpha)$ is trained with the squared-error loss at step $m$ to produce $\{h(\vx_i; \valpha_m)\}_1^N$ most parallel to $\{-g_m(\vx_i)\}_1^N$, and the solution $h(\vx;\valpha_m)$ can be applied to approximate $-g_m(\vx)$ for any $\vx$: \ba{\valpha_m=\argmin_{\vxi, \omega}\sum_{i=1}^N\big(-g_m(\vx_i)-\omega\cdot h(\vx_i;\vxi)\big)^2. \label{eq:grad_pred_obj}}

Therefore, with finite data, the gradient descent update in the function space at step $m$ is \ba{\hat{F}_m(\vx) = \hat{F}_{m-1}(\vx)+\rho_m\cdot h(\vx;\valpha_m), \label{eq:gradboost_one_step}} and the prediction of $\vy$ given any $\vx$ can be obtained through \ba{\hat{\vy}=\hat{F}^*(\vx) = \hat{F}_0(\vx)+\sum_{m=1}^M\rho_m\cdot h(\vx;\valpha_m). \label{eq:pred_gradboost}}

The function $h(\vx; \valpha)$ is termed a \textit{weak learner} or \textit{base learner}, and is often parameterized by a simple Classification And Regression Tree (CART) \citep{breiman84cart}. Eq.\,(\ref{eq:pred_gradboost}) has the form of an ensemble of weak learners, trained sequentially and combined via weighted sum. It is worth noting that when the loss function $L(\vy, F)$ is chosen to be the squared-error loss, its negative gradient is the residual: $-\frac{\partial L}{\partial F(\vx)}=\vy-F(\vx),$ and the optimal solution for minimizing this loss is the conditional mean, $\E[\vy\given\vx]$.

\subsection{Classification and Regression Diffusion Models (CARD)}
\label{ssec:CARD}
With the same goal as gradient boosting of tackling supervised learning problems, CARD \citep{han2022card} approaches them from a different angle: by adopting a generative modeling framework, a CARD model directly outputs samples from $p(\vy\given\vx)$, instead of summary statistics such as $\E[\vy\given\vx]$. This finer level of granularity in model output helps to paint a more complete picture of $p(\vy\given\vx)$. 
A unique advantage of CARD is that it does not require $p(\vy\given\vx)$ to adhere to a parametric form. 

At its core, CARD is a generative model that aims to learn a function parameterized by $\vtheta$ that maps a sample from a simple known distribution (\textit{i.e.}, the \textit{noise distribution}) to a sample from the target distribution $p(\vy\given\vx)$. As a generative model, its objective function is rooted in distribution matching: re-denoting the ground truth $p(\vy\given\vx)$ as $q(\vy_0\given\vx)$, we wish to learn $\vtheta$ so that $p_{\vtheta}(\vy_0\given\vx)$ approximates $q(\vy_0\given\vx)$ well, \textit{i.e.}, \ba{\kldiv[\big]{q(\vy_0\given\vx)}{p_{\vtheta}(\vy_0\given\vx)}\approx 0. \label{eq:kl_root_obj}} As a class of diffusion models, CARD produces a \textit{less noisy version} of $\vy$ after each function evaluation, which is then fed into the \textit{same} function to produce the next one. The final output $\vy_0$ can be viewed as a noiseless sample of $\vy$ from $p(\vy\given\vx)$. This autoregressive fashion of computing can be described as \textit{iterative refinement} or \textit{progressive denoising}.

The noisy samples of $\vy$ from the intermediate steps are treated as latent variables, linked together by a Markov chain with $T+1$ timesteps constructed in the direction \textit{opposite} to the data \textit{generation} process: with the stepwise transition distribution $q(\vy_t\given\vy_{t-1}, \vx)$, the \textit{forward diffusion process} is defined as $q(\vy_{1:T}\given\vy_0, \vx)=\prod_{t=1}^T q(\vy_t\given\vy_{t-1}, \vx)$. Meanwhile, the \textit{reverse diffusion process} is defined as $p_{\vtheta}(\vy_{0:T}\given\vx)=p(\vy_T\given\vx)\prod_{t=1}^T p_{\vtheta}(\vy_{t-1}\given\vy_t, \vx)$, in which $p(\vy_T\given\vx)=\gN(\vmu_T, \mI)$ is \textit{the} noise distribution, also referred to as the \textit{prior distribution}.

Utilizing the decomposition of cross entropy and Jensen's inequality, the variational bound (\textit{i.e.}, the negative ELBO) \citep{blei2017variational} can be derived from Eq.\,(\ref{eq:kl_root_obj}) as a new objective function, which can be further decomposed into terms at different timesteps \citep{jsd2015diffusion,ho2020ddpm}: \ba{L\eqqcolon\E_{q(\vy_{0:T}\given\vx)}\left[\log\frac{q(\vy_{1:T}\given\vy_0, \vx)}{p_{\vtheta}(\vy_{0:T}\given\vx)}\right]=\E_{q(\vy_{0:T}\given\vx)}\left[L_T+\sum_{t=2}^TL_{t-1}+L_0\right] \label{eq:card_neg_elbo_obj}.} It can be shown that the main focus for optimizing $\vtheta$ is on the $L_{t-1}$ terms for $t=2,\dots,T$, where \ba{L_{t-1} \coloneqq \kldiv[\big]{q(\vy_{t-1}\given\vy_t,\vy_0,\vx)}{p_{\vtheta}(\vy_{t-1}\given\vy_t,\vx)}.} An in-depth walkthrough of the objective function construction can be found in \cref{sssec:CARD}.

The distribution $q(\vy_{t-1}\given\vy_t,\vy_0,\vx)$ in each $L_{t-1}$ is called the \textit{forward process posterior distribution}, which is tractable and can be derived by applying Bayes' rule: 
\ba{q(\vy_{t-1}\given\vy_t, \vy_0, \vx) \propto q\big(\vy_t\given\vy_{t-1}, \vx\big)\cdot q\big(\vy_{t-1}\given\vy_0, \vx\big). \label{eq:card_post_via_conjugacy}} 
Both $q\big(\vy_t\given\vy_{t-1}, \vx\big)$ and $q\big(\vy_{t-1}\given\vy_0, \vx\big)$ in Eq.\,(\ref{eq:card_post_via_conjugacy}) are Gaussian: the former is the stepwise transition distribution in the forward process, defined as $q(\vy_t\given\vy_{t-1}, \vx) =\gN\big(\vy_t; \sqrt{\alpha_t}\vy_{t-1} + (1-\sqrt{\alpha_t})\vmu_T, \beta_t\mI\big)$, where $\beta_t$ is the $t$-th term of a predefined noise schedule $\beta_1,\dots,\beta_T$, and $\alpha_t \coloneqq 1-\beta_t$. This design gives rise to a closed-form distribution to sample $\vy_t$ at any arbitrary timestep $t$: 
\ba{q(\vy_t\given\vy_0, \vx) =\gN\big(\vy_t; \sqrt{\bar{\alpha}_t}\vy_0 + (1-\sqrt{\bar{\alpha}_t})\vmu_T, (1-\bar{\alpha}_t)\mI\big), \label{eq:card_arb_fwd_dist}}
in which $\bar{\alpha}_t \coloneqq \prod_{j=1}^t\alpha_j$. Each of the forward process posteriors thus has the form of 
\ba{q(\vy_{t-1}\given\vy_t, \vy_0, \vx)=\gN\Big(\vy_{t-1}; \tilde{\vmu}(\vy_t, \vy_0, \vmu_T), \tilde{\beta_t}\mI\Big), \label{eq:card_post_dist}}
where the variance $\tilde{\beta_t} \coloneqq \frac{1-\bar{\alpha}_{t-1}}{1-\bar{\alpha}_t}\beta_t$, and the mean \ba{\tilde{\vmu}(\vy_t, \vy_0, \vmu_T) \coloneqq \gamma_0\cdot\vy_0+\gamma_1\cdot\vy_t+\gamma_2\cdot\vmu_T, \label{eq:card_post_dist_mean}} in which the value of coefficients can be found in \cref{sssec:CARD}.

Now to minimize each $L_{t-1}$, $p_{\vtheta}(\vy_{t-1}\given\vy_t,\vx)$ needs to approximate the Gaussian distribution $q(\vy_{t-1}\given\vy_t,\vy_0,\vx)$, whose variance $\tilde{\beta_t}$ is already known. Therefore, the learning task is reduced to optimizing $\vtheta$ for the estimation of the forward process posterior mean $\tilde{\vmu}(\vy_t, \vy_0, \vmu_T)$. In other words, the data \textit{generation} process can now be modeled analytically, in the sense that an explicit distributional form (\textit{i.e.}, Gaussian) can be imposed upon adjacent latent variables. CARD adopts the noise-prediction loss introduced in \citet{ho2020ddpm}, a simplification of $L_{t-1}$: 
\ba{\gL_{\text{CARD}}=\E_{p(t,\vy_0\given\vx,\epsilonv)}\left[\big|\big|\epsilonv - \epsilonv_{\vtheta}\big(\vx, \vy_t, f_{\phi}(\vx), t\big)\big|\big|^2\right], \label{eq:noise_est_obj}}
in which $\epsilonv\sim\gN(\vzero, \mI)$ is sampled as the \textit{forward} process noise term, $\vy_t=\sqrt{\bar{\alpha}_t}\vy_0+(1-\sqrt{\bar{\alpha}_t})\vmu_T+\sqrt{1-\bar{\alpha}_t}\epsilonv$ is the sample from the \textit{forward} process distribution (\ref{eq:card_arb_fwd_dist}), and $f_{\phi}(\vx)$ is the point estimate of $\E[\vy\given\vx]$.

\section{The Diffusion Boosting Framework}
\label{sec:diff_boost}
Having established the objective functions of gradient boosting and CARD in \cref{sec:background}, we now proceed to discuss the connections between these two methods.

\subsection{Connections between Gradient Boosting and CARD}
\label{ssec:connections}
To begin with, we note that the functions in both methods can be viewed as \textbf{\textit{gradient estimators}}. For gradient boosting, each weak learner approximates the negative gradient of the objective at a particular optimization step (\ref{eq:grad_pred_obj}). Meanwhile, \citet{song2019scorematching} approach the training of diffusion models from the perspective of denoising score matching \citep{vincent2011connection}.

Specifically, in our supervised learning context, the conditional distribution of the noisy response variable given the covariates \(\vx\) can be modeled as a semi-implicit distribution \citep{yin2018semi,yu2023hierarchical}:
\ba{
q(\vy_t \given \vx) = \int q(\vy_t \given \vy_0, \vx) q(\vy_0 \given \vx) \, d\vy_0,
\label{eq:semi}
}
which generally lacks an analytic form since \(q(\vy_0 \given \vx)\) is unknown and is the target of our estimation from the observed \((\vx, \vy_0)\) pairs. This semi-implicit form allows for the estimation of its \textit{score}, the gradient of its log-likelihood with respect to the noisy sample that can be expressed as \(\nabla_{\vy_t} \log q(\vy_t \given \vx)\), using a score matching network \(\vs_{\vtheta}(\vx, \vy_t)\) \citep{zhou2024score}, as discussed below. 

Realizing score matching for supervised learning involves estimating the score by minimizing the explicit score matching (ESM) loss:
\ba{
\gL_{\text{ESM}}=\E_{t,\vy_0,\vy_t}\left[\lambda(t)||\nabla_{\vy_t}\log q(\vy_t\given\vx) - \vs_{\vtheta}(\vx,\vy_t)||^2\right],\label{eq:score_obj}
} 
where \(\lambda(t)\) is a positive weighting function. However, this objective function is intractable in practice since \(\nabla_{\vy_t} \log q(\vy_t \given \vx)\) is generally unknown. To address this issue, following the idea of denoising score matching (DSM) \citep{vincent2011connection,song2019scorematching}, this intractable objective can be rewritten into an equivalent~form:
\ba{
\gL_{\text{DSM}}=\E_{t,\vy_0,\vy_t}\left[\lambda(t)||\nabla_{\vy_t}\log q(\vy_t\given\vy_0,\vx) - \vs_{\vtheta}(\vx,\vy_t)||^2\right], \label{eq:score_est_obj}
} 
where $q(\vy_t\given\vy_0,\vx)$ is the forward sampling distribution whose gradient is analytic. 
With the Gaussian formulation of the forward process sampling distributions, Eqs.\,(\ref{eq:noise_est_obj}) and (\ref{eq:score_est_obj}) are connected via $\nabla_{\vy_t} \log %
q(\vy_t\given\vy_0,\vx)
=-\frac{\epsilonv}{\sqrt{1-\bar{\alpha}_t}}$, thus denoting $\epsilonv_{\vtheta}^{(t)}\coloneqq\epsilonv_{\vtheta}\big(\vx, \vy_t, f_{\phi}(\vx), t\big)$, we have $||\nabla_{\vy_t}\log %
q(\vy_t\given\vy_0,\vx)
- \vs_{\vtheta}(\vx,\vy_t)||^2=\frac{1}{1-\bar{\alpha}_t}||\epsilonv - \epsilonv_{\vtheta}^{(t)}||^2$. In other words, $\epsilonv_{\vtheta}^{(t)} \equiv -\sqrt{1-\bar{\alpha}_t} \cdot \vs_{\vtheta}(\vx,\vy_t)$  and  CARD estimates the (scaled) gradient of $\log
q(\vy_t\given\vx)$ at each diffusion timestep.

Additionally, we highlight that the core mechanism of both methods is \textbf{\textit{iterative refinement}}: gradient boosting essentially performs gradient descent in the function space (\cref{ssec:gradient_boosting}), and CARD generates each sample by making small and incremental changes to the initial noise sample over multiple steps to progressively refine it into a sample that resembles one from the target distribution (\cref{ssec:CARD}). The final form of gradient boosting is a strong function estimator (of a summary statistic), while CARD constructs a strong \textit{implicit} conditional distribution estimator.

Moreover, we point out that the iterative refinement mechanism implies \textbf{\textit{the adequacy of a weak learner}} at each refining step. This is already evident for gradient boosting, as each base learner is usually a single tree \citep{friedman2001gbm}. For CARD, we revisit the crucial term $L_{t-1}$ in the learning objective (\ref{eq:card_neg_elbo_obj}): in \cref{ssec:CARD}, we showed that the task of learning the diffusion model parameter $\vtheta$ is reframed as approximating the forward process posteriors $q(\vy_{t-1}\given\vy_t,\vy_0,\vx)$ with $p_{\vtheta}(\vy_{t-1}\given\vy_t,\vx)$ --- more specifically, estimating the mean term $\tilde{\vmu}(\vy_t, \vy_0, \vmu_T)$ with the diffusion model, since the variance can already be computed analytically. Note that Eq.\,(\ref{eq:card_post_dist_mean}) formulates $\tilde{\vmu}$ as a linear combination of the target variable $\vy_0$, the noisy sample $\vy_t$ from the previous timestep $t$, and the prior mean $\vmu_T$, with their corresponding coefficients $\gamma_0$, $\gamma_1$, and $\gamma_2$. During the data generation process, $\vy_0$ is unknown (and is the target that we seek to generate), thus at each timestep of this reverse process, CARD approximates this term via the reparameterization of the \textit{forward} process sampling distribution (\ref{eq:card_arb_fwd_dist}), in which the noise term is estimated by the noise-predicting network $\epsilonv_{\vtheta}$: \ba{\hat{\vy}_0=\frac{1}{\sqrt{\bar{\alpha}_t}}\Big(\vy_t-(1-\sqrt{\bar{\alpha}_t})\vmu_T-\sqrt{1-\bar{\alpha}_t}\epsilonv_{\vtheta}^{(t)}\Big). \label{eq:card_y0_fwd_reparam}} In other words, CARD generates new samples of $\vy$ by exploiting the Bayesian formulation of the reverse process stepwise transition distribution, \textit{i.e.}, the forward process posterior $q(\vy_{t-1}\given\vy_t, \vy_0, \vx)$: more specifically, CARD computes the \textit{surrogate} of the true $\vy_0$, providing the only missing piece in the analytical form of the posterior mean $\tilde{\vmu}$ to kickstart the sampling process.

\begin{figure}[t]
\setlength{\abovecaptionskip}{0pt} %
\setlength{\belowcaptionskip}{0pt} %
\begin{center}
\centerline{\includegraphics[width=0.5\textwidth]{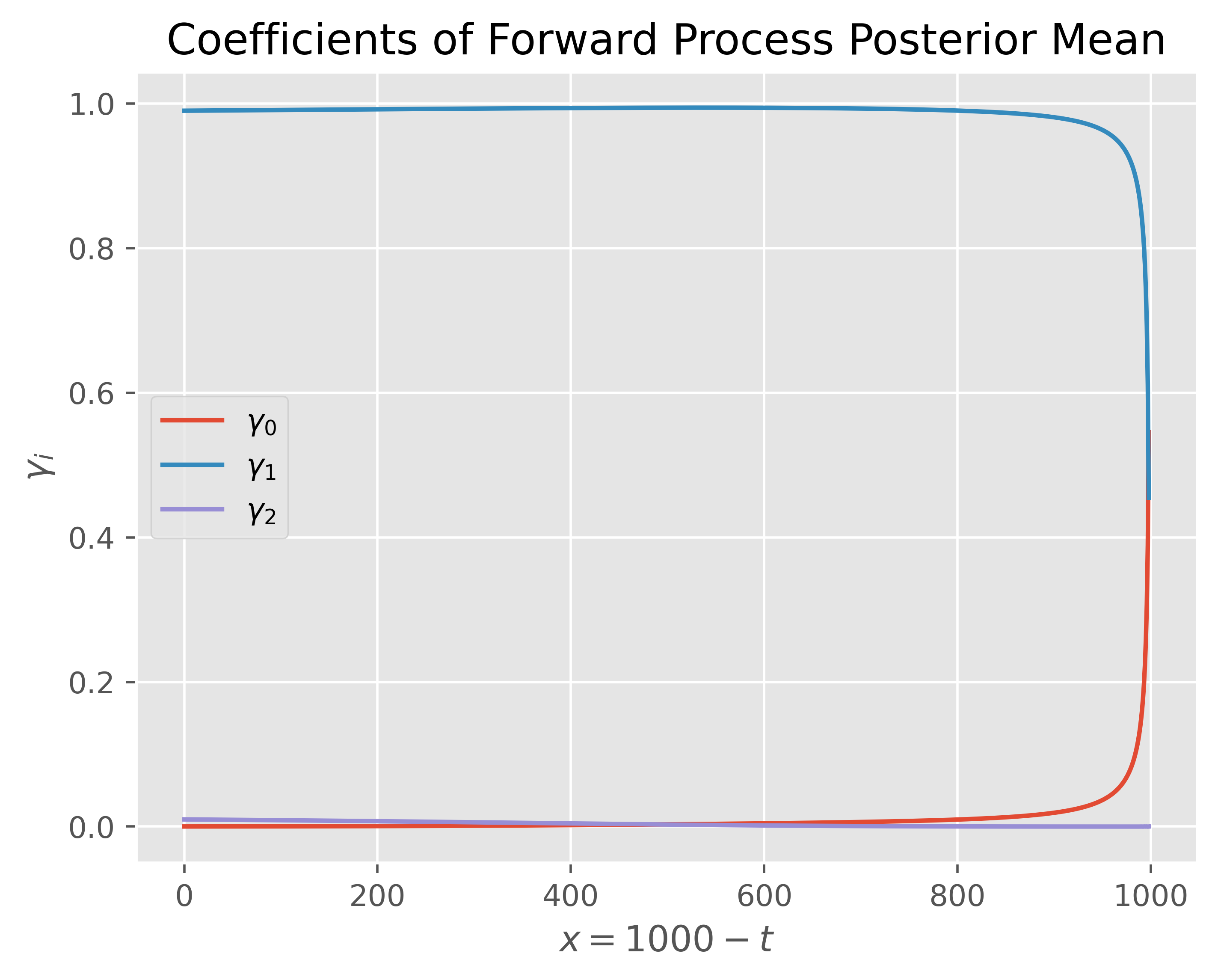}}

\caption{CARD posterior mean coefficients in Eq.\,(\ref{eq:card_post_dist_mean}) across all timesteps during sampling.}
\label{fig:fwd_post_mean_coef}
\end{center}
\vspace{-3mm}
\end{figure}

We take a closer look at the role of each term in the linear combination that forms the forward process posterior mean $\tilde{\vmu}$ (\ref{eq:card_post_dist_mean}), by plotting the coefficient values across all timesteps during the reverse process in \cref{fig:fwd_post_mean_coef}, where we set the total number of timesteps $T=1000$, and apply a linear noise schedule from $\beta_1=10^{-4}$ to $\beta_T=0.02$. The process starts at timestep $T$ (with label $0$ at the $x$-axis). Notice that $\gamma_2$ stays consistently close to $0$ for all timesteps, which makes intuitive sense, since the information of the prior mean $\vmu_T$ has been largely absorbed by the noise sample $\vy_T$. The more interesting part is the arcs of $\gamma_0$ and $\gamma_1$: across the vast majority of the timesteps (\textit{i.e.}, from $t=1000$ to around $t=100$), $\gamma_0$ stays very close to $0$, while $\gamma_1$ stays very close to $1$ --- this shows that prior to about the last $10\%$ of timesteps of the reverse process, the value of the posterior mean $\tilde{\vmu}$ is predominantly determined by $\vy_t$. In other words, the \textit{mean} of the next $\vy$ sample is almost the same as the \textit{value} of the current $\vy$ sample; at the same time, the contribution of $\hat{\vy}_0$ --- the surrogate of the true $\vy_0$ predicted by the diffusion model (\ref{eq:card_y0_fwd_reparam}) --- to the computation of $\tilde{\vmu}$ is basically negligible. The value of $\gamma_0$ begins to surge around the very end of the reverse process, by which time $\hat{\vy}_0$ shall be close enough to $\vy_0$ when the model is well-trained.

Based on the above observations regarding the computation of $\tilde{\vmu}$ --- specifically, that for most of the timesteps during sampling, the coefficient of $\vy_0$ is close to $0$, and during the remaining timesteps when $\hat{\vy}_0$ is close enough to $\vy_0$, the model only needs to capture small changes --- it is reasonable to argue that a weak learner at each timestep during the reverse diffusion process shall be sufficient in terms of computational power. In other words, we do not necessarily need a strong model to estimate $\vy_0$ at each timestep.

Having examined the similarities between gradient boosting and CARD, we now turn our attention to an analysis of their differences. More specifically, we come up with the following question.

\subsection{What can CARD learn from Gradient Boosting?}
\label{ssec:gb_improving_card}
In \cref{ssec:connections}, we established that for CARD, a weak learner should suffice to meet the computational requirements at each reverse timestep. This is because the noise-predicting network in CARD performs a similar task to each weak learner in gradient boosting, namely, approximating a gradient term. Moreover, the $\vy_0$ term, as part of the posterior mean computation, may not require a \textit{precise} estimation for most of the timesteps. However, these insights are not currently reflected in CARD's implementation: CARD uses a deep neural network to parameterize the noise predictor $\epsilonv_{\vtheta}$ due to the need for amortization, \textit{i.e.}, the same function is applied across all timesteps, necessitating it to possess abundant modeling and computational capacity. Therefore, based on our findings in \cref{ssec:connections}, it may be beneficial to follow the gradient boosting paradigm and improve the function choice of CARD by modeling the gradient term at each timestep with a \textit{different} weak learner.

Additionally, since CART \citep{breiman84cart} is the orthodox function choice for gradient boosting and tree-based models are widely acknowledged as superior to deep neural networks for tabular data \citep{grinsztajn2022outperform, qin2021outperform, tabdl2022, dltabsurvey2022, yang2018dndt}, incorporating CART into the CARD framework could potentially enhance CARD's ability to model tabular data, which represents the very type of data from which many supervised learning demands arise. Importantly, this new function choice could more clearly validate the DDPM framework \citep{ho2020ddpm} as a method whose success is grounded in statistical principles and computational practices (for example,  the Bayesian formulation with Gaussian conjugacy shown in Eq.\,(\ref{eq:card_post_via_conjugacy}), the variational lower bound, and reparameterization).
This perspective emphasizes the foundational statistical and computational mechanisms over the reliance on the architectural complexities of deep neural networks or the engineering nuances typically employed during training and sampling.

Furthermore, each weak learner in gradient boosting is trained sequentially (\ref{eq:gradboost_one_step}). Although the reverse process of CARD (and diffusion models in general) conducts sampling in a sequential fashion, its training treats the latent variables at different timesteps as independent random variables: during inference, $\vy_t$ is sampled from the approximation of the forward process posterior distribution (\ref{eq:card_post_dist}) (where the true $\vy_0$ is replaced with its surrogate), whose mean $\tilde{\vmu}$ depends on $\vy_{t+1}$, the sample from the previous timestep (\ref{eq:card_post_dist_mean}); however, during training, $\vy_t$ is drawn from the forward process sampling distribution (\ref{eq:card_arb_fwd_dist}). This creates a \textit{discrepancy} in the noisy response input $\vy_t$ for the $\epsilonv_{\vtheta}$ network \textit{between training and sampling}. This phenomenon of model input mismatch is commonly referred to as %
\textit{exposure bias} \citep{teacherforcing1989, bengio2015discrepancy, ranzato2016exposurebias, schmidt2019exposurebias, fan2020seqgen, ning2023expbias, ning2023reduceexpbias}. To address this issue, one could refer to the sequential training mechanism of gradient boosting (\ref{eq:grad_pred_obj}, \ref{eq:gradboost_one_step}) to devise a method for aligning the computational graphs during training and sampling.

\begin{algorithm}[t]
\begin{algorithmic}[1]
\REQUIRE Training set $\{(\vx_i, \vy_{0,i})\}_{i=1}^N$
\ENSURE Trained mean estimator $f_{\phi}(\vx)$ and tree ensemble $\{f_{\vtheta_{t}}\}_{t=1}^T$
\STATE Pre-train $f_{\phi}(\vx)$ to estimate $\E[\vy_0\given\vx]$
\FOR{$t=T$ to $1$}
\IF {$t = T$}
    \STATE Sample $\textcolor{blue}{\hat{\vy}_t}\sim\gN(\vmu_T, \mI)$, the prior distribution
\ELSE
    \STATE Sample $\vy_{t+1}\sim q(\vy_{t+1}\given\vy_0, \vx)$
    \STATE Predict $\vy_0$ with the newly trained model $f_{\vtheta_{t+1}}$: \bastar{\hat{\vy}_{0,t+1}=f_{\vtheta_{t+1}}\big(\vy_{t+1}, \vx, f_{\phi}(\vx)\big)}
    \STATE Compute $\tilde{\vmu}(\vy_{t+1}, \vy_0, \vmu_T)$, the forward process posterior mean: \bastar{\hat{\tilde{\vmu}}_t=\gamma_0\cdot\hat{\vy}_{0,t+1}+\gamma_1\cdot\vy_{t+1}+\gamma_2\cdot \vmu_T}
    \STATE Sample $\textcolor{blue}{\hat{\vy}_t}\sim\gN(\hat{\tilde{\vmu}}_t, \tilde{\beta}_{t+1}\mI)$ 
\ENDIF
\STATE Train $f_{\vtheta_{t}}$ with MSE loss to predict the true response: \bastar{\mathcal{L}_{\vtheta}^{(t)} = \E\left[\big|\big|\textcolor{red}{\vy_0} - f_{\vtheta_{t}}\big(\textcolor{blue}{\hat{\vy}_t}, \vx, f_{\phi}(\vx)\big)\big|\big|^2\right]}
\ENDFOR
\end{algorithmic}
\caption{Diffusion Boosted Trees Training}
\label{alg:dbt_train}
\end{algorithm}
\begin{algorithm}[th]
\begin{algorithmic}[1]
\REQUIRE Test data $\{\vx_j\}_{j=1}^M$, trained $f_{\phi}(\vx)$ and $\{f_{\vtheta_{t}}\}_{t=1}^T$
\ENSURE Response variable prediction $\hat{\vy}_{0,1}$
\STATE Draw $\hat{\vy}_T\sim\mathcal{N}(\vmu_T, \mI)$
\FOR{$t=T$ to $1$}
\STATE Predict the response $\hat{\vy}_{0,t}=f_{\vtheta_{t}}\big(\hat{\vy}_t, \vx, f_{\phi}(\vx)\big)$
\IF {$t > 1$}
    \STATE Draw the noisy sample $\hat{\vy}_{t-1}\sim q\big(\vy_{t-1}\given\hat{\vy}_t, \hat{\vy}_{0,t}, f_{\phi}(\vx)\big)$
\ENDIF
\ENDFOR
\STATE \textbf{return} $\hat{\vy}_{0,1}$
\end{algorithmic}
\caption{Diffusion Boosted Trees Sampling}
\label{alg:dbt_inference}
\end{algorithm}

\subsection{Diffusion Boosted Trees}
\label{ssec:dbm}
Following our discussion in \cref{ssec:gb_improving_card}, we now propose the \textbf{Diffusion Boosting} framework.

First, we replace the amortized single model $\epsilonv_{\vtheta}$ in the CARD framework with a series of weak learners $\{f_{\vtheta_{t}}\}_{t=1}^T$, \textit{one for each diffusion timestep}. For the input to each $f_{\vtheta_{t}}$, we use the same set of variables as CARD except the timestep $t$. Since we train a distinct model for each timestep, the representation of the temporal dynamic is no longer needed. We concatenate the remaining variables --- the noisy sample of $\vy$, the covariates $\vx$, and the conditional mean estimation $f_{\phi}(\vx)$ --- to form the model input. For simplicity, each $f_{\vtheta_{t}}$ directly predicts $\vy_0$ as its target, instead of the forward process noise sample $\epsilonv$ (\ref{eq:noise_est_obj}), thus sparing the step of converting the estimated $\epsilonv$ to $\hat{\vy}_0$ via Eq.\,(\ref{eq:card_y0_fwd_reparam}). 

We then choose CART \citep{breiman84cart} as the default function to parameterize each weak learner. For the inaugural algorithm, we set the number of trees to $1$ for each $f_{\vtheta_{t}}$, which is the universal setting applied to all the experiments in \cref{sec:experiments}. We argue that model performance could potentially be improved by using more trees when $\gamma_0$ (\ref{eq:card_post_dist_mean}) surges near the end of the generation process (\cref{fig:fwd_post_mean_coef}), \textit{i.e.}, when the estimate of $\vy_0$ has more impact on the computation of $\tilde{\vmu}$. We defer this attempt for future iterations of the algorithm.

Furthermore, to address the issue of exposure bias, we design a \textit{sequential training} paradigm inspired by gradient boosting: we train the first weak learner at timestep $T$, then use its output to construct the input for training the next weak learner at timestep $T-1$, and so on. This approach creates a dependency for adjacent weak learners \textit{during training}, emulating the computational graphs of consecutive timesteps during sampling.

Initially, we considered duplicating the sampling procedure during training, \textit{i.e.}, sampling a set of noises from the prior $\gN(\vmu_T, \mI)$ as the input to train $f_{\vtheta_{t}}$, then using the trained model with the same set of noise samples to predict $\hat{\vy}_{0,T}$, which would be used for the training of $f_{\vtheta_{T-1}}$, and so on. However, this would result in all $f_{\vtheta_{t}}$'s being trained with the same set of noise samples, limiting the diversity of training data and introducing additional overhead for storing the noisy samples during training. Therefore, we directly sample $\vy_{t+1}$ from $q(\vy_{t+1}\given\vy_0, \vx)$ (\ref{eq:card_arb_fwd_dist}), to be used as the input to the trained $f_{\vtheta_{t+1}}$ when training $f_{\vtheta_{t}}$.

Incorporating the above-mentioned modifications into CARD, we propose \textbf{D}iffusion \textbf{B}oosted \textbf{T}rees (DBT) as a class of diffusion boosting models. The training and sampling procedures are presented in Algorithms~\ref{alg:dbt_train}~and~\ref{alg:dbt_inference}, respectively, for both regression and classification tasks.

Notably, we made a slight adjustment to the CARD framework by writing the prior mean in the generic form $\vmu_T$ instead of $f_{\phi}(\vx)$. This introduces an extra degree of freedom by allowing the choice of the prior mean to differ from the conditional mean estimation $f_{\phi}(\vx)$, while still using $f_{\phi}(\vx)$ as an input to each $f_{\vtheta_{t}}$ since it possesses the information about $\E[\vy\given\vx]$.

The design choices and evaluation methods for diffusion boosting in both regression and classification tasks are presented as follows.

\subsubsection{Diffusion Boosting Regressor}
\label{sssec:dbm_regression}
For regression, the conditional mean estimator $f_{\phi}(\vx)$ is pre-trained with the MSE loss. It can be parameterized by any type of model, including neural networks, tree-based models, linear models with the OLS solution, \textit{etc.}

To evaluate a DBT model, we apply the conventional metrics \textbf{RMSE} and \textbf{NLL}, as well as the \textbf{QICE} metric proposed in \citet{han2022card}, which is a quantile-based coverage metric that measures the level of distribution matching between the true and the learned distributions. For data whose $\vx$ and $\vy$ are both 1D, a scatter plot can also be made for visual inspection of true and generated samples.

\subsubsection{Diffusion Boosting Classifier}
\label{sssec:dbm_classification}
For classification, we tailor the model to specifically tackle binary classification tasks on tabular data. This is a family of supervised learning tasks that CARD has not attempted, and it represents one of the most successful and common applications of tree-based models \citep{grinsztajn2022outperform}.

Unlike CARD, where the class representation of the response variable $\vy$ has the same dimensionality as the number of classes, we adopt a 1D representation of $\vy$. This choice is due to the fact that popular gradient boosting libraries do not natively support multi-dimensional outputs, and a scalar representation is sufficient for binary classification.

Binary classes are first encoded as scalar labels, $0$ and $1$, to pre-train $f_{\phi}(\vx)$ using the binary cross-entropy loss. This function outputs \textit{the predicted probability of the class with label $1$} to guide the training of DBT. These class labels are then converted to the logit scale to serve as class representations, also known as class prototypes in \citet{han2022card}, which have an unbounded range. This transformation aligns with the Gaussian assumption in the denoising diffusion framework, allowing us to use the same objective function to train DBT for both regression and classification.

We leverage the stochastic nature inherent in a generative model's output to evaluate DBT, following the paradigm proposed in \citet{han2022card}, with modifications for 1D output. For \textit{each test instance} $\vx_j$, we generate $S$ samples $\{\vy_{j,s}\}_{s=1}^S$ as class predictions in logits. The evaluation process consists of two steps:

\begin{enumerate}[topsep=0pt, leftmargin=*]
    \item Class Prediction:
    \begin{enumerate}[noitemsep, topsep=0pt, leftmargin=*]
        \item Apply the sigmoid function to convert each output to a probability, representing the predicted probability of label $1$: ${p_{j,s}}^{(1)}=\text{sigmoid}(\vy_{j,s})$.
        \item Convert these probabilities to binary labels using a threshold: $0.5$ for a balanced dataset, or the mean of the binary labels from the training set for an imbalanced one.
        \item Classify via the majority vote by the generated samples: selecting the more frequently predicted label as the class prediction.
    \end{enumerate}
    \item Model Confidence Measurement:
    \begin{enumerate}[noitemsep, topsep=0pt, leftmargin=*]
        \item \textbf{Prediction Interval Width (PIW)}: Compute the PIW between two percentile levels ($2.5^{th}$ and $97.5^{th}$ by default) of the $S$ samples (in either logit or probability). A narrower PIW indicates higher model confidence for that particular test instance, as it suggests less variation among the $S$ samples. Relative confidence can be assessed by comparing the PIWs of different test instances.
        \item \textbf{Paired Two-Sample $t$-Test}: Compute the corresponding class prediction of label $0$ for each of the $S$ samples: ${p_{j,s}}^{(0)}=1-{p_{j,s}}^{(1)}$. Perform a paired two-sample $t$-test to determine if $\{{p_{j,s}}^{(0)}\}_{s=1}^S$ and $\{{p_{j,s}}^{(1)}\}_{s=1}^S$ have significantly different sample means. Rejecting the $t$-test indicates the model is confident in its class prediction for $\vx_j$. The significance level, set to $\alpha=0.05$ by default, can be interpreted as an adjustable confidence level and can be modified based on the practical problem.
    \end{enumerate}
\end{enumerate}

\section{Related Work}
\label{sec:related_work}
Our work shares the same goal as CARD \citep{han2022card} to model the conditional distribution $p(\vy\given\vx)$ under a supervised learning setting from a generative modeling perspective, \textit{i.e.}, directly generating samples from $p(\vy\given\vx)$, rather than providing a point estimate. This method allows for the direct calculation of various summary statistics from the generated samples, including the conditional expectation $\E[\vy \given \vx]$, different quantiles such as the median, and measures of predictive uncertainty, thereby providing a more comprehensive representation of the target distribution. This generative method to capture $p(\vy\given\vx)$ has also been employed by \citet{jointmatching} and \citet{wganjointmatching}, both of which are based on GANs \citep{goodfellow2014gan} instead of diffusion models \citep{jsd2015diffusion} as the generative modeling framework. These works do not impose any parametric assumptions on the distributional form of $p(\vy\given\vx)$, allowing it to be learned completely from data.

The family of Bayesian neural networks (BNNs) \citep{bbb, mcdropout, pbp, alex2017uncertainty, vdropout, concretedropout} is another class of methods that is capable of capturing predictive uncertainty. Unlike our work and CARD, which focus exclusively on modeling aleatoric uncertainty, BNNs address both aleatoric and epistemic uncertainty \citep{aneuncertainty2021} by treating network parameters as random variables. Furthermore, BNNs often explicitly assume $p(\vy\given\vx)$ to be Gaussian, facilitating the decomposition of these two types of predictive uncertainty \citep{uncertaintydecomp2018}. 
Another line of work related to ours applies a semi-implicit construction \citep{yin2018semi}, $\int p(\vy\given \vx,\vz)p(\vz\given \vx)\mathrm d\vz$, to model local uncertainties \citep{wang2020thompson}. In this approach, local variables are typically infused with uncertainty through contextual dropout \citep{fan2021cdropout,boluki2020learnable}, while auto-encoding variational inference \citep{kingma2014vae} is employed to obtain point estimates of the underlying neural networks.

Ensemble-based methods also assess predictive uncertainty by assuming a Gaussian form for $p(\vy\given\vx)$, attaining aleatoric uncertainty by learning both the mean and the variance parameter via the Gaussian negative log-likelihood objective, and quantifying epistemic uncertainty by training multiple base models. It can be parameterized by either deep neural networks \citep{deepensembles} or trees \citep{ngboost2020, gbdtuncertainty2021}. Bayesian Additive Regression Trees (BART) \citep{chipman2010, bartrpackage2021, he2019, starling2020, hill2011} is another class of ensemble methods, which approximates $\E[\vy\given\vx]$ by a sum of regression trees. It assumes the conventional additive Gaussian noise regression model, and obtains MCMC samples of the sum-of-trees model and the noise variance parameter from their corresponding posterior distributions. It can be viewed as a Bayesian form of gradient boosting.

Additionally, several studies feature components akin to our work. Forest-Diffusion \citep{ajm2023forestdiffusion} is the first work to parameterize diffusion-based generative models with gradient boosted trees (GBTs), and is designed for unconditional tabular data generation and imputation. This approach involves training a distinct GBT model at each diffusion timestep, with each model comprising 100 trees. eDiff-I \citep{balaji2022ediffi} approaches text-to-image generation by training an ensemble of three expert denoisers, each specialized for a specific timestep interval, instead of using a single model across all timesteps. This method aims to capture the complex temporal dynamics observed at different stages of generation: throughout the process, the dependence of the denoising model gradually shifts from the input text prompt embedding towards the visual features. SGLB \citep{sglb2021} introduces a gradient boosting algorithm that leverages the Langevin diffusion equation to facilitate convergence to the global optimum during numerical optimization, regardless of the loss function's convexity.

\section{Experiments}
\label{sec:experiments}
Our current implementation of DBT is based on the LightGBM \citep{ke2017lightgbm} library. For each $f_{\vtheta_{t}}$, we fix the number of trees at $1$, and set the default number of leaves to $101$ and the learning rate to $1$, leaving all other hyperparameters at their default settings. Since LightGBM requires loading the entire dataset for training, we need to construct the training data in its entirety, instead of iteratively updating the model via mini-batches. We set the number of noise samples for each instance $n_{noise}=100$, and duplicate the entire dataset $n_{noise}$ times to construct the training set. To address the inefficiency of duplicating the training set, we plan to incorporate the XGBoost \citep{chen2016xgboost} library in future iterations of our code. Recent versions of XGBoost offer the data iterator functionality for memory-efficient training with external memory, eliminating the need to duplicate the training set, as demonstrated by \citet{ajm2023forestdiffusion} in their updated Forest-Diffusion repository.

The input to each $f_{\vtheta_{t}}$, $\big(\hat{\vy}_t, \vx, f_{\phi}(\vx)\big)$, is formed via concatenation. The conditional mean estimator $f_{\phi}(\vx)$ is conceptually model-free. When training on complete data, we use the same parameterization as CARD: a feedforward neural network with two hidden layers, containing $100$ and $50$ hidden units, respectively. A Leaky ReLU activation function with a $0.01$ negative slope is employed after each hidden layer. For datasets with missing covariates, we parameterize $f_{\phi}(\vx)$ using a gradient-boosted trees model. This model consists of $100$ trees with $31$ leaf nodes each, and it is trained with a learning rate of $0.05$.

For the diffusion model hyperparameters, we set the number of timesteps $T=1000$, and use a linear noise schedule with $\beta_1=10^{-4}$ and $\beta_T=0.02$. While we currently apply the same set of hyperparameters for $f_{\vtheta_{t}}$ across all timesteps, each tree is free to be trained with different hyperparameter settings, achieving a new level of flexibility over an amortized deep neural network. We reserve the exploration in this direction for future work.

\subsection{Regression}
\label{ssec:reg}
For regression, we incorporate experiments on both toy and real-world datasets.

\subsubsection{Toy Examples}
\label{sssec:reg_toy}
We designed several toy examples with diverse statistical attributes --- including linear and non-linear piecewise-defined functions with additive Gaussian noise, multimodality, and heteroscedasticity --- to demonstrate the following: 1) like CARD, DBT is versatile in modeling conditional distributions; 2) DBT is better suited for functions where the response variable, $\vy$, exhibits distinct, non-continuous values across subintervals of $\vx$; and 3) DBT requires less data than CARD to reach effective performance levels.

\begin{figure}[htbp]
\vspace{-2mm}
\begin{center}
\centerline{\includegraphics[width=0.92\textwidth]{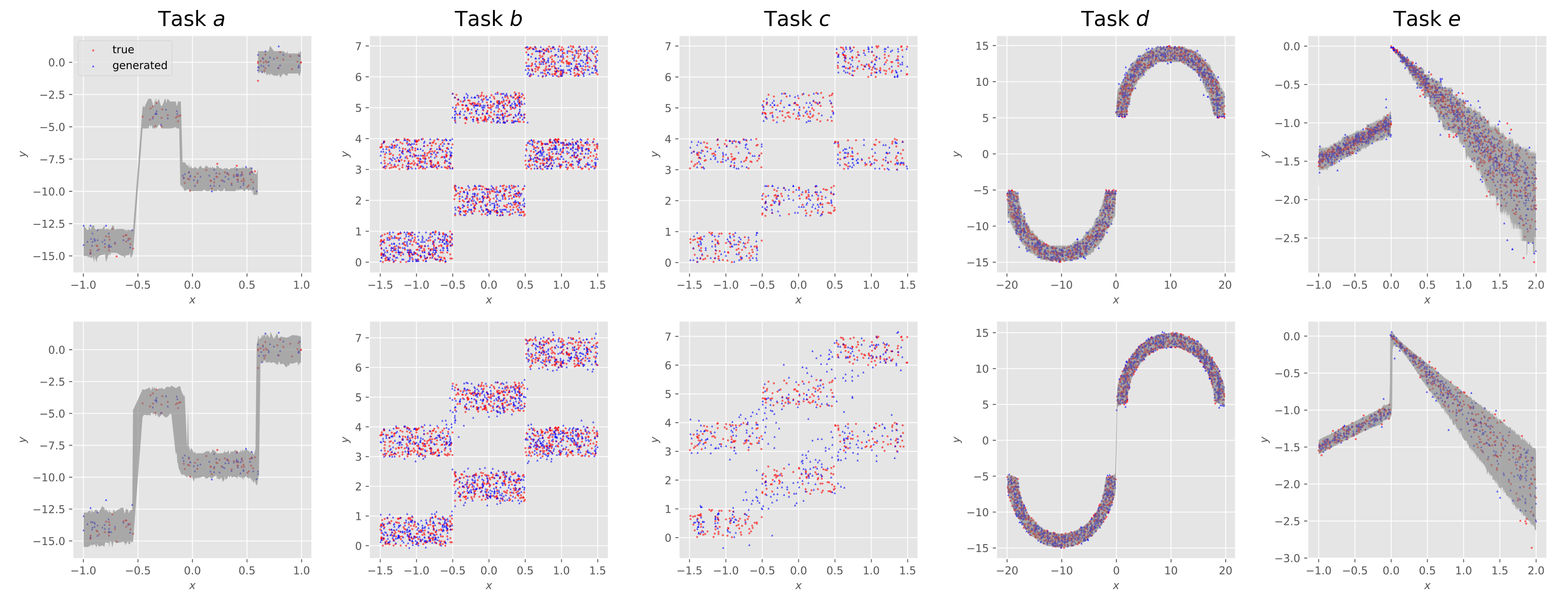}}
\caption{Comparison of DBT (\textbf{top row}) and CARD (\textbf{bottom row}) on toy regression examples.}
\label{fig:dbm_vs_card_toy}
\end{center}
\vspace{-5mm}
\end{figure}

We train both DBT and CARD on these datasets and create scatter plots with both true and generated samples, as illustrated in \cref{fig:dbm_vs_card_toy}. The tasks are denoted from left to right as \textit{a}, \textit{b}, \textit{c}, \textit{d}, and \textit{e}. For tasks \textit{a}, \textit{d}, and \textit{e}, where $\vy$ is unimodal, we shade the region between the $2.5^{th}$ and $97.5^{th}$ percentiles of the generated $\vy$ in grey. Tasks~\textit{b} and \textit{c} are based on the same true data-generating function; however, Task~\textit{c} utilizes only $\tfrac{1}{5}$ of the training data compared to Task~\textit{b}.

We observe that DBT consistently generates samples that blend very well with the true data across all tasks, highlighting its capability to accurately capture the underlying data generation mechanisms.

Furthermore, for the uni-modal tasks \textit{a}, \textit{d}, and \textit{e}, there is a notable distinction in the central $95\%$ sample intervals between DBT and CARD. Specifically, in areas near the junctions of two adjacent $\vx$ subintervals, CARD tends to create a visible ``band'' that bridges these subintervals. In contrast, DBT forms either a much narrower stripe or no stripe at all, more effectively capturing the ``disjointness'' of $\vy$. This observation underscores a clear advantage of trees over deep neural networks as a function choice: trees make predictions by dividing the covariate space into discrete subregions, making them naturally suited for modeling piecewise-defined functions. Conversely, deep neural networks, which are smooth functions due to the use of activation functions, tend to interpolate between breakpoints in adjacent $\vx$ subintervals by ``borrowing'' information from the different $\vy$ values in these neighborhoods.

Lastly, for the multimodal tasks \textit{b} and \textit{c}, the consequences of the different function choices between DBT and CARD are amplified. While DBT generates samples with clear-cut boxes, CARD struggles to separate these regions in the generated samples. This is particularly evident in Task~\textit{c}: with only one-fifth the training data compared to Task~\textit{b}, DBT still produces samples that align closely with the true data, whereas the samples generated by CARD blend together, failing to separate $\vy$ in each of the three $\vx$ subintervals, revealing the challenges CARD faces in scenarios with limited data availability.

\subsubsection{OpenML Examples}
\label{sssec:reg_openml}
We now turn our attention to real-world datasets that exhibit discontinuities in the feature space rather than the response variable space. We conduct an ablation study to demonstrate the impact of the three key adjustments made to CARD for the construction of DBT (\cref{ssec:gb_improving_card}), namely: switching from one single amortized model to a series of weak learners, tree parameterization, and sequential training.

Initially, we planned to compare the performance of DBT and CARD along with two variants of CARD: one where the amortized neural network is replaced with an amortized GBT, and another in which the amortized neural network is replaced with distinct single tree models at different timesteps, trained independently rather than sequentially. The former model was eliminated from the list due to its consistently poor performance on UCI \citep{ucidataset} benchmark datasets (\cref{ssec:amortized_gbt_ablation_study}). The latter model demonstrated reasonable results and was retained. This model is named CARD-T --- refer to \cref{ssec:card_t_algo} for details on CARD-T's training and sampling algorithms.

We identified two benchmark datasets characterized by a large number of categorical features from \citet{grinsztajn2022outperform}, available on OpenML \citep{OpenML2013}: \textit{Mercedes\_Benz\_Greener\_Manufacturing}, which contains $4,209$ instances with all $359$ features being categorical, and \textit{Allstate\_Claims\_Severity}, which comprises $188,318$ datapoints, with $110$ out of $124$ features being categorical.

For both datasets, we trained the models on $5$ different train-test splits, each with a $90\%/10\%$ ratio, created using distinct random seeds. Table~\ref{tab:reg_openml_metrics} presents the results evaluated with RMSE, NLL, and QICE (\cref{sssec:dbm_regression}). Our observations indicate that both DBT and CARD-T significantly outperform CARD in terms of distribution matching, as reflected by lower NLL and QICE values, while also providing better mean estimates. Furthermore, DBT surpasses CARD-T in all but one instance, highlighting the effectiveness of the sequential training scheme in enhancing the tree-based model's performance.

\begin{figure}[htbp]
\vspace{-1mm}
\centering
\centering
\captionof{table}{\label{tab:reg_openml_metrics}OpenML regression tasks.}
\vspace{-3mm}
\resizebox{0.45\textwidth}{!}{
\begin{tabular}{@{}l|ccc@{}}
\toprule[1.5pt]
\multirow{2}{*}{Dataset} & DBT & CARD-T & CARD \\ \cmidrule{2-4}
& \multicolumn{3}{c}{RMSE $\downarrow$} \\ \midrule
Mercedes & $\bm{8.19\pm 1.50}$ & $8.37\pm 1.75$ & $8.80\pm 0.88$ \\
Allstate & $\bm{0.55\pm 0.00}$ & $0.56\pm 0.00$ & $0.60\pm 0.00$ \\ \midrule\midrule[1.5pt]
& \multicolumn{3}{c}{NLL $\downarrow$} \\ \midrule
Mercedes & $\bm{3.40\pm 0.04}$ & $3.52\pm 0.05$ & $7.85\pm 1.81$ \\
Allstate & $\bm{0.93\pm 0.00}$ & $1.19\pm 0.00$ & $1.07\pm 0.03$ \\ \midrule\midrule[1.5pt]
& \multicolumn{3}{c}{QICE $\downarrow$} \\ \midrule
Mercedes & $\bm{1.11\pm 0.36}$ & $1.35\pm 0.27$ & $6.36\pm 0.32$ \\
Allstate & $0.34\pm 0.05$ & $\bm{0.25\pm 0.07}$ & $3.18\pm 0.10$ \\ \bottomrule\bottomrule[1.5pt]
\end{tabular}
}
\vspace{10mm}
\centering
\captionof{table}{\label{tab:reg_uci_metric_tables}UCI regression tasks.}
\resizebox{\textwidth}{!}{
\begin{tabular}{@{}l|ccccccc@{}}
\toprule[1.5pt]
\multirow{2}{*}{Dataset} & PBP & MC Dropout & Deep Ensembles & GCDS & CARD & \textbf{DBT} & \textbf{DBT ($10\%$ MCAR)} \\ \cmidrule{2-8}
& \multicolumn{7}{c}{RMSE $\downarrow$} \\ \midrule
Boston & $2.89\pm 0.74$ & $3.06\pm 0.96$ & $3.17\pm 1.05$ & $2.75\pm 0.58$ & $\bm{2.61\pm 0.63}$ & $\bm{2.73\pm 0.62}$ & $3.30\pm 0.89$ \\
Concrete & $5.55\pm 0.46$ & $5.09\pm 0.60$ & $4.91\pm 0.47$ & $5.39\pm 0.55$ & $\bm{4.77\pm 0.46}$ & $\bm{4.56\pm 0.50}$ & $5.17\pm 0.58$ \\
Energy & $1.58\pm 0.21$ & $1.70\pm 0.22$ & $2.02\pm 0.32$ & $0.64\pm 0.09$ & $\bm{0.52\pm 0.07}$ & $\bm{0.52\pm 0.07}$ & $0.62\pm 0.14$ \\
Kin8nm & $9.42\pm 0.29$ & $7.10\pm 0.26$ & $8.65\pm 0.47$ & $8.88\pm 0.42$ & $\bm{6.32\pm 0.18}$ & $\bm{7.04\pm 0.23}$ & $13.60\pm 0.46$ \\
Naval & $0.41\pm 0.08$ & $0.08\pm 0.03$ & $0.09\pm 0.01$ & $0.14\pm 0.05$ & $\bm{0.02\pm 0.00}$ & $\bm{0.07\pm 0.01}$ & $0.25\pm 0.01$ \\
Power & $4.10\pm 0.15$ & $4.04\pm 0.14$ & $4.02\pm 0.15$ & $4.11\pm 0.16$ & $\bm{3.93\pm 0.17}$ & $3.95\pm 0.16$ & $\bm{3.72\pm 0.16}$ \\
Protein & $4.65\pm 0.02$ & $4.16\pm 0.12$ & $4.45\pm 0.02$ & $4.50\pm 0.02$ & $\bm{3.73\pm 0.01}$ & $\bm{3.81\pm 0.04}$ & $4.35\pm 0.04$ \\
Wine & $0.64\pm 0.04$ & $\bm{0.62\pm 0.04}$ & $0.63\pm 0.04$ & $0.66\pm 0.04$ & $0.63\pm 0.04$ & $\bm{0.61\pm 0.04}$ & $0.65\pm 0.04$ \\
Yacht & $0.88\pm 0.22$ & $0.84\pm 0.27$ & $1.19\pm 0.49$ & $\bm{0.79\pm 0.26}$ & $\bm{0.65\pm 0.25}$ & $1.08\pm 0.39$ & $1.12\pm 0.34$ \\
Year & $8.86\pm$ NA & $\bm{8.77\pm}$ NA & $8.79\pm$ NA & $9.20\pm$ NA & $\bm{8.70\pm}$ NA & $8.81\pm$ NA & $9.23\pm$ NA \\ \midrule
\# Top 2 & $0$ & $2$ & $0$ & $1$ & $\bm{9}$ & $\bm{7}$ & $1$ \\ \midrule\midrule[1.5pt]
& \multicolumn{7}{c}{NLL $\downarrow$} \\ \midrule
Boston & $2.53\pm 0.27$  & $2.46\pm 0.12$ & $2.35\pm 0.16$ & $18.66\pm 8.92$ & $\bm{2.35\pm 0.12}$ & $\bm{2.33\pm 0.12}$ & $3.77\pm 1.34$ \\
Concrete & $3.19\pm 0.05$  & $3.21\pm 0.18$ & $\bm{2.93\pm 0.12}$  & $13.64\pm 6.88$ & $2.96\pm 0.09$ & $\bm{2.91\pm 0.08}$ & $3.06\pm 0.09$ \\
Energy & $2.05\pm 0.05$ & $1.50\pm 0.11$ & $1.40\pm 0.27$ & $1.46\pm 0.72$ & $\bm{1.04\pm 0.06}$ & $\bm{0.91\pm 0.21}$ & $3.58\pm 10.69$ \\
Kin8nm & $-0.83\pm 0.02$ & $\bm{-1.14\pm 0.05}$ & $-1.06\pm 0.02$ & $-0.38\pm 0.36$ & $\bm{-1.32\pm 0.02}$ & $-1.09\pm 0.02$ & $-0.56\pm 0.02$ \\
Naval & $-3.97\pm 0.10$  & $-4.45\pm 0.38$ & $\bm{-5.94\pm 0.10}$ & $-5.06\pm 0.48$ & $\bm{-7.54\pm 0.05}$ & $-4.31\pm 0.05$ & $-3.84\pm 0.02$ \\
Power & $2.92\pm 0.02$  & $2.90\pm 0.03$ & $2.89\pm 0.02$ & $2.83\pm 0.06$ & $\bm{2.82\pm 0.02}$ & $2.90\pm 0.02$ & $\bm{2.81\pm 0.02}$ \\
Protein & $3.05\pm 0.00$  & $2.80\pm 0.08$ & $2.89\pm 0.02$ & $2.81\pm 0.09$  & $\bm{2.49\pm 0.03}$ & $\bm{2.64\pm 0.02}$ & $2.78\pm 0.02$ \\
Wine & $1.03\pm 0.03$  & $0.93\pm 0.06$ & $0.96\pm 0.06$ & $6.52\pm 21.86$ & $\bm{0.92\pm 0.05}$ & $\bm{0.88\pm 0.04}$ & $13.94\pm 10.06$ \\
Yacht & $1.58\pm 0.08$  & $1.73\pm 0.22$ & $1.11\pm 0.18$ & $\bm{0.61\pm 0.34}$  & $0.90\pm 0.08$ & $\bm{0.59\pm 0.24}$ & $0.91\pm 0.20$ \\
Year & $3.69\pm$ NA & $\bm{3.42\pm}$ NA & $3.44\pm$ NA & $3.43\pm$ NA & $\bm{3.34\pm}$ NA & $3.44\pm$ NA & $3.52\pm$ NA \\ \midrule
\# Top 2 & $0$ & $2$ & $2$ & $1$ & $\bm{8}$ & $\bm{6}$ & $1$ \\ \midrule\midrule[1.5pt]
& \multicolumn{7}{c}{QICE (in $\%$) $\downarrow$} \\ \midrule
Boston & $3.50\pm 0.88$ & $3.82\pm 0.82$ & $\bm{3.37\pm 0.00}$ & $11.73\pm 1.05$ & $\bm{3.45\pm 0.83}$ & $4.19\pm 1.18$ & $9.26\pm 1.36$ \\
Concrete & $2.52\pm 0.60$ & $4.17\pm 1.06$ & $2.68\pm 0.64$ & $10.49\pm 1.01$ & $\bm{2.30\pm 0.66}$ & $\bm{2.52\pm 0.51}$ & $4.21\pm 0.92$ \\
Energy & $6.54\pm 0.90$ & $5.22\pm 1.02$ & $\bm{3.62\pm 0.58}$ & $7.41\pm 2.19$ & $4.91\pm 0.94$ & $\bm{3.78\pm 0.91}$ & $5.48\pm 1.15$ \\
Kin8nm & $1.31\pm 0.25$ & $1.50\pm 0.32$ & $\bm{1.17\pm 0.22}$ & $7.73\pm 0.80$ & $\bm{0.92\pm 0.25}$ & $1.31\pm 0.29$ & $1.30\pm 0.27$ \\
Naval & $4.06\pm 1.25$ & $12.50\pm 1.95$ & $6.64\pm 0.60$ & $5.76\pm 2.25$ & $\bm{0.80\pm 0.21}$ & $11.16\pm 1.66$ & $\bm{1.78\pm 0.26}$ \\
Power & $\bm{0.82\pm 0.19}$  & $1.32\pm 0.37$ & $1.09\pm 0.26$ & $1.77\pm 0.33$ & $0.92\pm 0.21$ & $1.24\pm 0.20$ & $\bm{0.91\pm 0.23}$ \\
Protein & $1.69\pm 0.09$ & $2.82\pm 0.41$ & $2.17\pm 0.16$ & $2.33\pm 0.18$ & $\bm{0.71\pm 0.11}$ & $0.95\pm 0.10$ & $\bm{0.73\pm 0.18}$ \\
Wine & $\bm{2.22\pm 0.64}$ & $2.79\pm 0.56$ & $\bm{2.37\pm 0.63}$ & $3.13\pm 0.79$ & $3.39\pm 0.69$ & $8.18\pm 1.17$ & $13.91\pm 0.58$ \\
Yacht & $6.93\pm 1.74$ & $10.33\pm 1.34$ & $7.22\pm 1.41$ & $\bm{5.01\pm 1.02}$ & $8.03\pm 1.17$ & $\bm{5.96\pm 1.51}$ & $6.26\pm 1.53$ \\
Year & $2.96\pm$ NA & $2.43\pm$ NA & $2.56\pm$ NA & $1.61\pm$ NA & $\bm{0.53\pm}$ NA & $1.07\pm$ NA & $\bm{0.72\pm}$ NA \\ \midrule
\# Top 2 & $2$ & $0$ & $\bm{4}$ & $1$ & $\bm{6}$ & $3$ & $4$ \\ \bottomrule\bottomrule[1.5pt]
\end{tabular}
}
\vspace{3mm}
\end{figure}

\subsubsection{SHAP Value Analysis on OpenML Dataset}
\label{sssec:reg_shap}
One major strength of decision tree-based models over neural networks is their interpretability: they provide clear visualizations of decision paths and the influence of each feature on the outcome. By employing distinct models for each diffusion timestep, rather than a single amortized model for all timesteps, we are able to develop a unique method to investigate the impact of each input feature on the prediction at different diffusion timesteps.

\begin{figure}[htbp]
\begin{center}
\centerline{\includegraphics[width=1.0\textwidth]{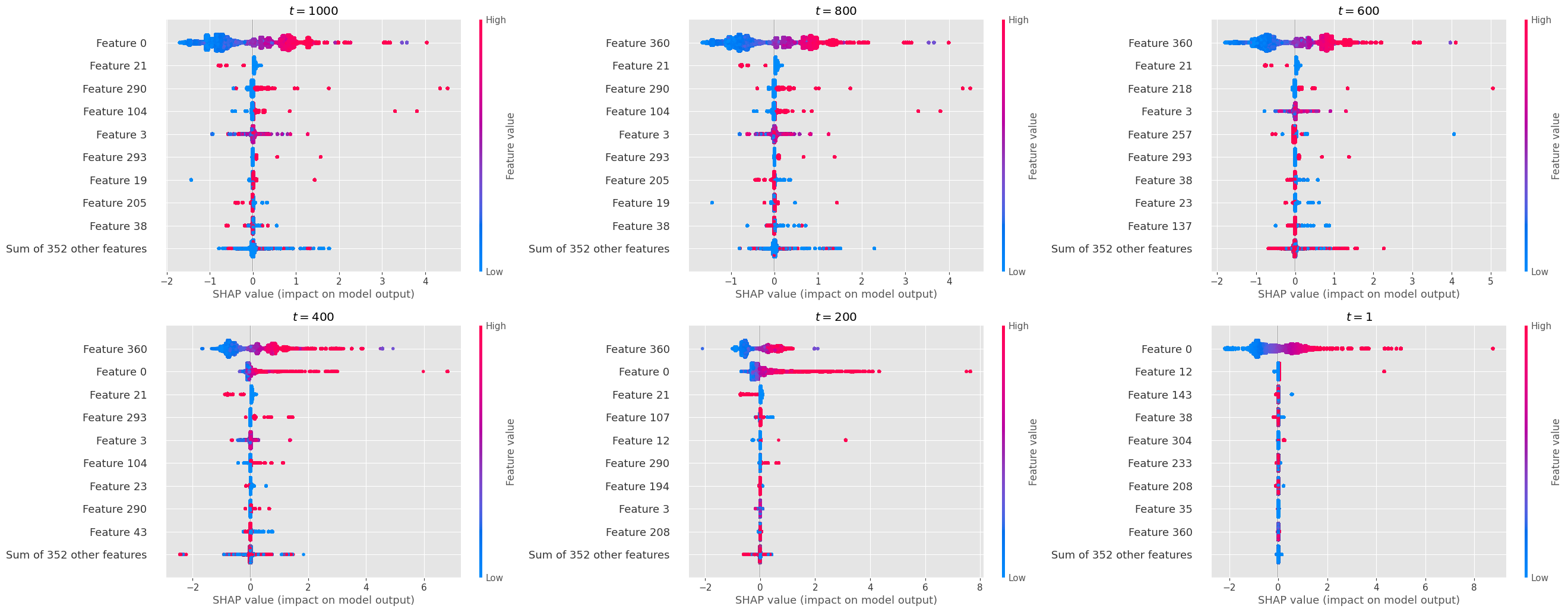}}
\caption{Beeswarm summary plots of SHAP values at six diffusion timesteps.}
\label{fig:openml_shap}
\end{center}
\end{figure}

Using DBT's trained models {\footnotesize $\{f_{\vtheta_{t}}\}_{t=1}^T$} on the \textit{Mercedes} dataset (Table~\ref{tab:reg_openml_metrics}), we generated beeswarm summary plots of SHAP values \citep{shap2017} at six timesteps: $t=1000, 800, 600, 400, 200, 1$, as shown in \cref{fig:openml_shap}. In each plot, the features are sorted by their magnitude of impact on model output, measured by the sum of SHAP values over all training samples.

The input to each $f_{\vtheta_{t}}$ is the concatenated vector $\big(\hat{\vy}_t, \vx, f_{\phi}(\vx)\big)$. Since $\vx$ is $359$-dimensional, ``Feature 0'' represents the noisy sample $\hat{\vy}_t$, and ``Feature 360'' is the estimation of $\E[\vy\given\vx]$, $f_{\phi}(\vx)$. We observe that ``Feature 360'' remains the most impactful feature at the four intermediate timesteps ($t=800, 600, 400, 200$), highlighting the pivotal role of the pre-trained mean estimator $f_{\phi}(\vx)$ in guiding the sampling process. For the final model during sampling, $f_{\vtheta_{1}}$, the most influential feature has changed to ``Feature 0'': the output is almost solely affected by the sample $\hat{\vy}_1$, which should be very close to the true $\vy_0$ if the model is well-trained.

Additionally, we observe changes in the ranking of other important features, indicating that the relative impact of each feature on the model's predictions varies over the course of generation.

We also provide a set of feature importance plots at the same six timesteps in \cref{fig:openml_feature_importance}, along with an analysis comparing SHAP values and feature importance, in \cref{ssec:feature_importance}.

\subsubsection{UCI Datasets}
\label{sssec:reg_uci}
We apply the same paradigm as \citet{han2022card} to benchmark DBT on real-world regression datasets. A detailed description of the experimental setup can be found in \cref{ssec:uci_reg_experiment_setup}. In addition to training DBT on the full dataset, we also evaluate its performance on incomplete data, where $10\%$ of the covariate values are randomly removed (Missing Completely at Random; MCAR). The evaluation metrics are reported in Table~\ref{tab:reg_uci_metric_tables}, along with the number of times each model achieves a Top-2 ranking based on each metric.

Our observations show that DBT trained on the full dataset achieves performance on par with CARD, while still outperforming other baseline methods. This demonstrates the effectiveness of our proposed method in modeling conditional distributions in real-world settings. It is important to note that all our experiments are based on inherently heterogeneous tabular data, characterized by variability in feature types, as well as in the number of features and samples. Therefore, introducing a new method for modeling such data should aim to provide a competitive alternative among state-of-the-art approaches, offering practitioners an additional tool for tackling new datasets. More discussion on this point can be found in \cref{ssec:performance_on_uci_reg}.

A distinct advantage of DBT over CARD becomes apparent when dealing with data containing missing values. DBT handles missing data without the need for imputation and demonstrates robust performance: it rarely records the worst metric when compared to other baseline models trained on complete data, occasionally achieves state-of-the-art results in terms of RMSE and NLL, and performs well in terms of QICE. This robustness makes DBT particularly useful in real-world applications where missing data is prevalent, such as healthcare, finance, and survey analysis, providing a reliable and efficient solution for handling incomplete tabular datasets.

\subsection{Classification}
\label{ssec:clf}
For classification, we contextualize the diffusion boosting framework through a practical business application: credit card fraud detection.

\subsubsection{The Story of OTA Fraud Detection}
\label{sssec:ota_fraud_detection}
As generative AI has been advancing at a staggering speed in the past few years in terms of the fidelity of content generation, its negative social impact has also become increasingly concerning \citep{kenthapadi2023genai, grace2024aifuture}. Against this backdrop, this section endeavors to pivot the discourse towards the beneficial potential of generative AI: we introduce a fraud detection paradigm that leverages the stochasticity of generative models as its foundational mechanism. Our aim is to illuminate one positive application of generative AI, demonstrating its potential to contribute significantly to societal well-being.

We introduce the use case of fraud detection as a business component within an online travel agency (OTA), where transactions, including payments and bookings, are conducted digitally. This environment places OTAs at risk of credit card fraud, since it is hard to verify that the credit card user is indeed its rightful owner. Traditionally, companies have relied on rule-based models and human agents to identify suspicious transactions. However, the vast daily volume of transactions that came with complicated fraud patterns, coupled with the significant costs associated with employing a large team of agents, renders this approach impractical for scrutinizing every transaction.

In response to these challenges, OTA companies have gradually integrated machine learning techniques into their fraud detection ecosystem in recent years. These methods involve deploying classifiers that assess each transaction in real-time, and pass the dubious ones to human agents for further review. This hybrid approach significantly alleviates the burden on human agents by automating the initial screening process.

This operational model exemplifies a strategy known as \textit{learning to defer (L2D)} \citep{madras2018l2d, hari2022l2d, verma2022cl2d}, where AI systems recognize \textit{when} to rely on human expertise for decision-making, thus providing a more efficient workflow while ensuring more reliable outcomes. This idea is pivotal in the contexts where AI's \textit{decision confidence} is crucial.

\subsubsection{Binary Classification on Real-World Tabular Data}
\label{sssec:clf_openml}
We demonstrate how DBT conducts binary classification on tabular data using the evaluation framework described in \cref{sssec:dbm_classification}. We train the DBT model on a credit card default dataset \citep{yeh2009creditcarddefault}, which is another benchmark dataset from \citet{grinsztajn2022outperform}. This dataset contains $21$ covariates with both numerical and categorical features. The model is trained on $11,944$ instances and evaluated on the remaining $1,328$ cases.

A pre-trained neural network classifier $f_{\phi}(\vx)$ predicts the test set with an accuracy of $57.68\%$. For evaluation, we generate only 10 samples for each test instance. Firstly, we make predictions using the majority-voted label, achieving an improved accuracy of $69.58\%$. We then compute the PIW for all test instances and summarize the results in \cref{tab:piw_by_pred_class_table}. We observe that for the group of test instances predicted as Class $1$, higher accuracy is accompanied by a narrower mean PIW. Furthermore, within each predicted class, instances with correct predictions have a narrower mean PIW compared to those with incorrect predictions. Additionally, we group the test instances by increasing PIW values and compute the accuracy within each bin, as shown in \cref{tab:ranked_piw_table}\footnote{The bins are sorted in ascending order of PIW. There are only four distinct PIW values, since a tree model has a limited number of possible outputs (\textit{i.e.}, the number of leaves). Note that Bin 2 has a smaller PIW than Bin 3, although this is not reflected due to rounding to two decimal places.}. We observe that as the mean PIW increases from Bin 1 to Bin 4, the accuracy consistently decreases. The results from these two tables suggest that less variation in generated samples is associated with better performance in classification.

\begin{figure}[h]
\centering
\small
\captionof{table}{\label{tab:piw_by_pred_class_table}PIW for both majority-vote predicted class labels.}
\vspace{-3mm}
\begin{tabular}{@{}c|c|c|c|c@{}}
\toprule[1.05pt] \hline
\multirow{2}{*}{predicted class} & \multirow{2}{*}{accuracy} & \multicolumn{3}{c}{mean PIW} \\ \cline{3-5} %
& & overall (count) & correct pred. (count) & incorrect pred. (count) \\ \hline\hline %
0 & $66.14\%$ & $110.03$ $(762)$ & $108.10$ $(504)$ & $113.79$ $(258)$ \\ \hline
1 & $74.20\%$ & $86.50$ $(566)$ & $79.77$ $(420)$ & $105.86$ $(146)$ \\ \hline
\bottomrule[1.05pt]
\end{tabular}
\captionof{table}{\label{tab:ranked_piw_table}Accuracy across different PIW bins.}
\begin{tabular}{@{}c|c|c|c|c@{}}
\toprule[1.05pt]
Bin & 1 & 2 & 3 & 4 \\ \hline
mean PIW & $0.00$ & $94.44$ & $94.44$ & $121.86$ \\ \hline
accuracy (count) & $87.36\%$ $(182)$ & $82.31\%$ $(130)$ & $70.00\%$ $(120)$ & $64.06\%$ $(896)$ \\ 
\bottomrule[1.05pt]
\end{tabular}
\captionof{table}{\label{tab:overall_ttest_table}Accuracy by $t$-test outcomes.}
\begin{tabular}{@{}c|c|c@{}}
\toprule[1.05pt]
\multirow{2}{*}{$t$-test outcome} & \multicolumn{2}{c}{accuracy (count)} \\ \cmidrule{2-3}
& $\alpha=0.05$ & $\alpha=0.005$ \\ \hline\hline
reject & $77.79\%$ $(707)$ & $81.02\%$ $(432)$ \\ \hline
fail to reject & $60.23\%$ $(621)$ & $64.06\%$ $(896)$ \\ 
\bottomrule[1.05pt]
\end{tabular}
\captionof{table}{\label{tab:ttest_by_pred_class_table}Accuracy by predicted class labels.}
\resizebox{1\textwidth}{!}{
\begin{tabular}{@{}c|c||c|c|c|c|c|c@{}}
\toprule[1.05pt] \hline
\multirow{3}{*}{predicted class} & \multirow{3}{*}{accuracy} & \multicolumn{3}{c|}{$\alpha=0.05$} & \multicolumn{3}{c}{$\alpha=0.005$} \\ \cline{3-8} %
& & \multirow{2}{*}{$t$-test reject rate} & \multicolumn{2}{c|}{accuracy} & \multirow{2}{*}{$t$-test reject rate} & \multicolumn{2}{c}{accuracy} \\ \cline{4-5}\cline{7-8} %
& & & reject (count) & fail to reject (count) & & reject (count) & fail to reject (count) \\ \hline\hline %
0 & $66.14\%$ & $43.96\%$ & $71.94\%$ $(335)$ & $61.59\%$ $(427)$ & $21.92\%$ & $73.05\%$ $(167)$ & $64.20\%$ $(595)$ \\ \hline
1 & $74.20\%$ & $65.72\%$ & $83.06\%$ $(372)$ & $57.22\%$ $(194)$ & $46.82\%$ & $86.04\%$ $(265)$ & $63.79\%$ $(301)$ \\ \hline
\bottomrule[1.05pt]
\end{tabular}
}
\end{figure}

We now conduct the $t$-test on each test instance at two significance levels, $0.05$ and $0.005$. For each significance level, we observe in \cref{tab:overall_ttest_table} that the accuracy for test instances with rejected $t$-tests is considerably higher than for those where the $t$-tests fail to reject. This observation holds at each predicted class level, as shown in \cref{tab:ttest_by_pred_class_table}.

By comparing the accuracy and $t$-test reject rates between the two predicted classes, we further conclude that the more accurate class exhibits a higher rate of $t$-test null hypotheses being rejected. This validates the $t$-test as an effective method for measuring model confidence.

This evaluation design aligns seamlessly with the requirements of a learning-to-defer method: we can interpret cases where the $t$-tests fail to reject as uncertain predictions made by the DBT model, which can then be deferred to human agents for further evaluation. Assuming human agents can achieve the same level of accuracy as the cases with rejected $t$-tests, we can improve the overall accuracy from $69.58\%$ to $76.68\%$ with $\alpha=0.05$, and to $78.59\%$ with $\alpha=0.005$.

Furthermore, by comparing the results between the two significance levels in both Tables \ref{tab:overall_ttest_table} and \ref{tab:ttest_by_pred_class_table}, we observe the role of the significance level as a measure of decision conservativeness: a lower significance level implies a more conservative decision strategy, resulting in fewer instances where the $t$-tests are rejected, \textit{i.e.}, fewer predictions are made with confidence.

\section{Conclusion}
\label{sec:conclusion}
We propose the \textbf{Diffusion Boosting} paradigm as a new supervised learning algorithm, combining the merits of both \textit{Classification and Regression Diffusion Models (CARD)} and \textit{Gradient Boosting}. We implement \textbf{Diffusion Boosted Trees (DBT)}, which parameterizes the diffusion model by a single tree at each timestep. We demonstrate through experiments the advantages of DBT over CARD, and present a case study of fraud detection for DBT to perform classification on tabular data with the ability of learning to defer.

\section*{Acknowledgments}
The authors acknowledge the Texas Advanced Computing Center (TACC) for providing HPC and storage resources that have contributed to the research results reported within this paper. The authors would also like to thank Ruijiang Gao and Huangjie Zheng for their discussions during the course of this project.

\newpage
\bibliography{references}

\begin{thebibliography}{71}
\providecommand{\natexlab}[1]{#1}
\providecommand{\url}[1]{\texttt{#1}}
\expandafter\ifx\csname urlstyle\endcsname\relax
  \providecommand{\doi}[1]{doi: #1}\else
  \providecommand{\doi}{doi: \begingroup \urlstyle{rm}\Url}\fi

\bibitem[Balaji et~al.(2022)Balaji, Nah, Huang, Vahdat, Song, Zhang, Kreis, Aittala, Aila, Laine, Catanzaro, Karras, and Liu]{balaji2022ediffi}
Yogesh Balaji, Seungjun Nah, Xun Huang, Arash Vahdat, Jiaming Song, Qinsheng Zhang, Karsten Kreis, Miika Aittala, Timo Aila, Samuli Laine, Bryan Catanzaro, Tero Karras, and Ming-Yu Liu.
\newblock e{D}iff-{I}: Text-to-image diffusion models with an ensemble of expert denoisers.
\newblock \emph{ArXiv}, abs/2211.01324, 2022.

\bibitem[Bengio et~al.(2015)Bengio, Vinyals, Jaitly, and Shazeer]{bengio2015discrepancy}
Samy Bengio, Oriol Vinyals, Navdeep Jaitly, and Noam Shazeer.
\newblock Scheduled sampling for sequence prediction with recurrent neural networks.
\newblock In \emph{Proceedings of the 29th Conference on Neural Information Processing Systems}, 2015.

\bibitem[Blei et~al.(2017)Blei, Kucukelbir, and McAuliffe]{blei2017variational}
David~M Blei, Alp Kucukelbir, and Jon~D McAuliffe.
\newblock {V}ariational {I}nference: A review for statisticians.
\newblock \emph{Journal of the American Statistical Association}, 112\penalty0 (518):\penalty0 859--877, 2017.

\bibitem[Blundell et~al.(2015)Blundell, Cornebise, Kavukcuoglu, and Wierstra]{bbb}
Charles Blundell, Julien Cornebise, Koray Kavukcuoglu, and Daan Wierstra.
\newblock Weight uncertainty in neural network.
\newblock In \emph{Proceedings of the 32nd International Conference on Machine Learning}. PMLR, 2015.

\bibitem[Boluki et~al.(2020)Boluki, Ardywibowo, Dadaneh, Zhou, and Qian]{boluki2020learnable}
Shahin Boluki, Randy Ardywibowo, Siamak~Zamani Dadaneh, Mingyuan Zhou, and Xiaoning Qian.
\newblock Learnable {B}ernoulli dropout for {B}ayesian deep learning.
\newblock In \emph{Proceedings of the 23th International Conference on Artificial Intelligence and Statistics}, volume 108, pages 3905--3916. PMLR, 2020.

\bibitem[Borisov et~al.(2022)Borisov, Leemann, Se{\ss}ler, Haug, Pawelczyk, and Kasneci]{dltabsurvey2022}
Vadim Borisov, Tobias Leemann, Kathrin Se{\ss}ler, Johannes Haug, Martin Pawelczyk, and Gjergji Kasneci.
\newblock Deep neural networks and tabular data: A survey.
\newblock \emph{IEEE Transactions on Neural Networks and Learning Systems}, 99:\penalty0 1--21, 2022.

\bibitem[Breiman et~al.(1984)Breiman, Friedman, Stone, and Olshen]{breiman84cart}
Leo Breiman, Jerome Friedman, Charles~J. Stone, and R.A. Olshen.
\newblock \emph{Classification and Regression Trees}.
\newblock Chapman and Hall/CRC, 1984.

\bibitem[Cauchy(1847)]{cauchy1847}
A.~Cauchy.
\newblock M\'ethode g\'en\'erale pour la r\'esolution des syst\`emes d'\'equations simultan\'ees.
\newblock \emph{Comptes rendus de l'Acad\'emie des Sciences}, 25:\penalty0 536--538, 1847.

\bibitem[Chen and Guestrin(2016)]{chen2016xgboost}
Tianqi Chen and Carlos Guestrin.
\newblock {XGBoost}: A scalable tree boosting system.
\newblock In \emph{Proceedings of the 22nd ACM SIGKDD International Conference on Knowledge Discovery and Data Mining}, KDD '16, pages 785--794, 2016.

\bibitem[Chipman et~al.(2010)Chipman, George, and McCulloch]{chipman2010}
Hugh~A. Chipman, Edward~I. George, and Robert~E. McCulloch.
\newblock {BART}: {B}ayesian {A}dditive {R}egression {T}rees.
\newblock \emph{The Annals of Applied Statistics}, 4\penalty0 (1):\penalty0 266–298, 2010.

\bibitem[Depeweg et~al.(2018)Depeweg, Hernández-Lobato, Doshi-Velez, and Udluft]{uncertaintydecomp2018}
Stefan Depeweg, José~Miguel Hernández-Lobato, Finale Doshi-Velez, and Steffen Udluft.
\newblock Decomposition of uncertainty in bayesian deep learning for efficient and risk-sensitive learning.
\newblock In \emph{Proceedings of the 35th International Conference on Machine Learning}. PMLR, 2018.

\bibitem[Dhariwal and Nichol(2021)]{dhariwal2021diffbeatsgan}
Prafulla Dhariwal and Alexander~Quinn Nichol.
\newblock Diffusion models beat {GAN}s on image synthesis.
\newblock In \emph{Proceedings of the 35th Conference on Neural Information Processing Systems}, 2021.

\bibitem[Dua and Graff(2017)]{ucidataset}
Dheeru Dua and Casey Graff.
\newblock {UCI} {M}achine {L}earning {R}epository, 2017.
\newblock URL \url{http://archive.ics.uci.edu/ml}.

\bibitem[Duan et~al.(2020)Duan, Avati, Ding, Thai, Basu, Ng, and Schuler]{ngboost2020}
Tony Duan, Anand Avati, Daisy~Yi Ding, Khanh~K. Thai, Sanjay Basu, Andrew~Y. Ng, and Alejandro Schuler.
\newblock {NGB}oost: Natural gradient boosting for probabilistic prediction.
\newblock In \emph{Proceedings of the 37th International Conference on Machine Learning}. PMLR, 2020.

\bibitem[Fan et~al.(2020)Fan, Zhang, Wang, and Zhou]{fan2020seqgen}
Xinjie Fan, Yizhe Zhang, Zhendong Wang, and Mingyuan Zhou.
\newblock Adaptive correlated monte carlo for contextual categorical sequence generation.
\newblock In \emph{Proceedings of the 8th International Conference on Learning Representations}, 2020.

\bibitem[Fan et~al.(2021)Fan, Zhang, Tanwisuth, Qian, and Zhou]{fan2021cdropout}
Xinjie Fan, Shujian Zhang, Korawat Tanwisuth, Xiaoning Qian, and Mingyuan Zhou.
\newblock Contextual dropout: An efficient sample-dependent dropout module.
\newblock In \emph{Proceedings of the 9th International Conference on Learning Representations}, 2021.

\bibitem[Friedman(2001)]{friedman2001gbm}
Jerome~H. Friedman.
\newblock {G}reedy {F}unction {A}pproximation: {A} {G}radient {B}oosting {M}achine.
\newblock \emph{The Annals of Statistics}, 29\penalty0 (5):\penalty0 1189--1232, 2001.

\bibitem[Gal and Ghahramani(2016)]{mcdropout}
Yarin Gal and Zoubin Ghahramani.
\newblock Dropout as a {B}ayesian approximation: Representing model uncertainty in deep learning.
\newblock In \emph{Proceedings of the 33rd International Conference on Machine Learning}. PMLR, 2016.

\bibitem[Gal et~al.(2017)Gal, Hron, and Kendall]{concretedropout}
Yarin Gal, Jiri Hron, and Alex Kendall.
\newblock Concrete dropout.
\newblock In \emph{Proceedings of the 31st Conference on Neural Information Processing Systems}, 2017.

\bibitem[Goodfellow et~al.(2014)Goodfellow, Pouget-Abadie, Mirza, Xu, Warde-Farley, Ozair, Courville, and Bengio]{goodfellow2014gan}
Ian Goodfellow, Jean Pouget-Abadie, Mehdi Mirza, Bing Xu, David Warde-Farley, Sherjil Ozair, Aaron Courville, and Yoshua Bengio.
\newblock {G}enerative {A}dversarial {N}ets.
\newblock In \emph{Proceedings of the 27th Conference on Neural Information Processing Systems}, 2014.

\bibitem[Grace et~al.(2024)Grace, Stewart, Sandk{\"u}hler, Thomas, Weinstein-Raun, and Brauner]{grace2024aifuture}
Katja Grace, Harlan Stewart, Julia~Fabienne Sandk{\"u}hler, Stephen Thomas, Ben Weinstein-Raun, and Jan Brauner.
\newblock Thousands of {AI} authors on the future of {AI}.
\newblock \emph{ArXiv}, abs/2401.02843, 2024.

\bibitem[Grinsztajn et~al.(2022)Grinsztajn, Oyallon, and Varoquaux]{grinsztajn2022outperform}
L{\'e}o Grinsztajn, Edouard Oyallon, and Ga{\"e}l Varoquaux.
\newblock Why do tree-based models still outperform deep learning on tabular data?
\newblock In \emph{Proceedings of the 36th Conference on Neural Information Processing Systems}, 2022.

\bibitem[Han et~al.(2022)Han, Zheng, and Zhou]{han2022card}
Xizewen Han, Huangjie Zheng, and Mingyuan Zhou.
\newblock {CARD}: {C}lassification and {R}egression {D}iffusion {M}odels.
\newblock In \emph{Proceedings of the 36th Conference on Neural Information Processing Systems}, 2022.

\bibitem[He et~al.(2019)He, Yalov, and Hahn]{he2019}
Jingyu He, Saar Yalov, and P.~Richard Hahn.
\newblock {XBART}: {A}ccelerated {B}ayesian {A}dditive {R}egression {T}rees.
\newblock In \emph{Proceedings of the 22nd International Conference on Artificial Intelligence and Statistics}, volume~89, pages 1130--1138. PMLR, 2019.

\bibitem[He et~al.(2015)He, Zhang, Ren, and Sun]{he2015resnet}
Kaiming He, X.~Zhang, Shaoqing Ren, and Jian Sun.
\newblock Deep residual learning for image recognition.
\newblock \emph{2016 IEEE Conference on Computer Vision and Pattern Recognition (CVPR)}, pages 770--778, 2015.

\bibitem[Hernández-Lobato and Adams(2015)]{pbp}
José~Miguel Hernández-Lobato and Ryan~P. Adams.
\newblock Probabilistic backpropagation for scalable learning of {B}ayesian neural networks.
\newblock In \emph{Proceedings of the 32nd International Conference on Machine Learning}. PMLR, 2015.

\bibitem[Hill(2011)]{hill2011}
Jennifer~L. Hill.
\newblock Bayesian nonparametric modeling for causal inference.
\newblock \emph{Journal of Computational and Graphical Statistics}, 20\penalty0 (1):\penalty0 217--240, 2011.

\bibitem[Ho et~al.(2020)Ho, Jain, and Abbeel]{ho2020ddpm}
Jonathan Ho, Ajay Jain, and Pieter Abbeel.
\newblock Denoising diffusion probabilistic models.
\newblock In \emph{Proceedings of the 34th Conference on Neural Information Processing Systems}, 2020.

\bibitem[Hornik et~al.(1989)Hornik, Stinchcombe, and White]{hornik1989ffnnunivapprox}
Kurt Hornik, Maxwell Stinchcombe, and Halbert White.
\newblock Multilayer feedforward networks are universal approximators.
\newblock \emph{Neural Networks}, 2\penalty0 (5):\penalty0 359--366, 1989.

\bibitem[H\"ullermeier and Waegeman(2021)]{aneuncertainty2021}
Eyke H\"ullermeier and Willem Waegeman.
\newblock {A}leatoric and {E}pistemic {U}ncertainty in {M}achine {L}earning: An introduction to concepts and methods.
\newblock \emph{Machine Learning}, 110:\penalty0 457–506, 2021.

\bibitem[Jolicoeur-Martineau et~al.(2024)Jolicoeur-Martineau, Fatras, and Kachman]{ajm2023forestdiffusion}
Alexia Jolicoeur-Martineau, Kilian Fatras, and Tal Kachman.
\newblock Generating and imputing tabular data via diffusion and flow-based gradient-boosted trees.
\newblock In \emph{Proceedings of the 27th International Conference on Artificial Intelligence and Statistics}, 2024.

\bibitem[Karras et~al.(2022)Karras, Aittala, Aila, and Laine]{karras2022edm}
Tero Karras, Miika Aittala, Timo Aila, and Samuli Laine.
\newblock Elucidating the design space of diffusion-based generative models.
\newblock In \emph{Proceedings of the 36th Conference on Neural Information Processing Systems}, 2022.

\bibitem[Ke et~al.(2017)Ke, Meng, Finley, Wang, Chen, Ma, Ye, and Liu]{ke2017lightgbm}
Guolin Ke, Qi~Meng, Thomas Finley, Taifeng Wang, Wei Chen, Weidong Ma, Qiwei Ye, and Tie-Yan Liu.
\newblock {LightGBM}: A highly efficient gradient boosting decision tree.
\newblock In \emph{Proceedings of the 31st International Conference on Machine Learning}, 2017.

\bibitem[Kendall and Gal(2017)]{alex2017uncertainty}
Alex Kendall and Yarin Gal.
\newblock What uncertainties do we need in {B}ayesian deep learning for computer vision?
\newblock In \emph{Proceedings of the 31st Conference on Neural Information Processing Systems}, 2017.

\bibitem[Kenthapadi et~al.(2023)Kenthapadi, Lakkaraju, and Rajani]{kenthapadi2023genai}
Krishnaram Kenthapadi, Himabindu Lakkaraju, and Nazneen Rajani.
\newblock Generative {AI} meets responsible {AI}: Practical challenges and opportunities.
\newblock \emph{Proceedings of the 29th ACM SIGKDD Conference on Knowledge Discovery and Data Mining}, 2023.

\bibitem[Kingma and Welling(2014)]{kingma2014vae}
Diederik~P. Kingma and Max Welling.
\newblock {A}uto-{E}ncoding {V}ariational {B}ayes.
\newblock In \emph{Proceedings of the 2nd International Conference on Learning Representations}, 2014.

\bibitem[Kingma et~al.(2015)Kingma, Salimans, and Welling]{vdropout}
Durk~P. Kingma, Tim Salimans, and Max Welling.
\newblock Variational dropout and the local reparameterization trick.
\newblock In \emph{Proceedings of the 29th Conference on Neural Information Processing Systems}, 2015.

\bibitem[Lakshminarayanan et~al.(2017)Lakshminarayanan, Pritzel, and Blundell]{deepensembles}
Balaji Lakshminarayanan, Alexander Pritzel, and Charles Blundell.
\newblock Simple and scalable predictive uncertainty estimation using deep ensembles.
\newblock In \emph{Proceedings of the 31st Conference on Neural Information Processing Systems}, 2017.

\bibitem[Liu et~al.(2021)Liu, Zhou, Jiao, and Huang]{wganjointmatching}
Shiao Liu, Xingyu Zhou, Yuling Jiao, and Jian Huang.
\newblock Wasserstein generative learning of conditional distribution.
\newblock \emph{arXiv}, abs/2112.10039, 2021.

\bibitem[Lundberg and Lee(2017)]{shap2017}
Scott Lundberg and Su-In Lee.
\newblock A unified approach to interpreting model predictions.
\newblock In \emph{Proceedings of the 31st Conference on Neural Information Processing Systems}, 2017.

\bibitem[Madras et~al.(2018)Madras, Pitassi, and Zemel]{madras2018l2d}
David Madras, Toniann Pitassi, and Richard~S. Zemel.
\newblock Predict responsibly: Improving fairness and accuracy by learning to defer.
\newblock In \emph{Proceedings of the 32nd Conference on Neural Information Processing Systems}, 2018.

\bibitem[Malinin et~al.(2021)Malinin, Prokhorenkova, and Ustimenko]{gbdtuncertainty2021}
Andrey Malinin, Liudmila Prokhorenkova, and Aleksei Ustimenko.
\newblock Uncertainty in gradient boosting via ensembles.
\newblock In \emph{Proceedings of the 9th International Conference on Learning Representations}, 2021.

\bibitem[Narasimhan et~al.(2022)Narasimhan, Jitkrittum, Menon, Rawat, and Kumar]{hari2022l2d}
Harikrishna Narasimhan, Wittawat Jitkrittum, Aditya~K. Menon, Ankit Rawat, and Sanjiv Kumar.
\newblock Post-hoc estimators for learning to defer to an expert.
\newblock In \emph{Proceedings of the 36th Conference on Neural Information Processing Systems}, 2022.

\bibitem[Ning et~al.(2023{\natexlab{a}})Ning, Li, Su, Salah, and Ertugrul]{ning2023expbias}
Mang Ning, Mingxiao Li, Jianlin Su, A.~A. Salah, and Itir~Onal Ertugrul.
\newblock Elucidating the exposure bias in diffusion models.
\newblock \emph{arXiv}, abs/2308.15321, 2023{\natexlab{a}}.

\bibitem[Ning et~al.(2023{\natexlab{b}})Ning, Sangineto, Porrello, Calderara, and Cucchiara]{ning2023reduceexpbias}
Mang Ning, E.~Sangineto, Angelo Porrello, Simone Calderara, and Rita Cucchiara.
\newblock Input perturbation reduces exposure bias in diffusion models.
\newblock In \emph{Proceedings of the 40th International Conference on Machine Learning}. PMLR, 2023{\natexlab{b}}.

\bibitem[Nisan and Szegedy(1994)]{nisan1994boolfunc}
Noam Nisan and Mario Szegedy.
\newblock On the degree of boolean functions as real polynomials.
\newblock \emph{Computational Complexity}, 4:\penalty0 301–313, 1994.

\bibitem[Nocedal and Wright(1999)]{nocedalwright1999numopt}
Jorge Nocedal and Stephen~J. Wright.
\newblock Line search methods.
\newblock In \emph{Numerical Optimization}, chapter~3. Springer, 1999.

\bibitem[Prokhorenkova et~al.(2018)Prokhorenkova, Gusev, Vorobev, Dorogush, and Gulin]{lp2018catboost}
Liudmila Prokhorenkova, Gleb Gusev, Aleksandr Vorobev, Anna~Veronika Dorogush, and Andrey Gulin.
\newblock {CatBoost}: unbiased boosting with categorical features.
\newblock In \emph{Proceedings of the 32nd Conference on Neural Information Processing Systems}, 2018.

\bibitem[Qin et~al.(2021)Qin, Yan, Zhuang, Tay, Pasumarthi, Wang, Bendersky, and Najork]{qin2021outperform}
Zhen Qin, Le~Yan, Honglei Zhuang, Yi~Tay, Rama~Kumar Pasumarthi, Xuanhui Wang, Michael Bendersky, and Marc Najork.
\newblock Are neural rankers still outperformed by gradient boosted decision trees?
\newblock In \emph{Proceedings of the 9th International Conference on Learning Representations}, 2021.

\bibitem[Ranzato et~al.(2016)Ranzato, Chopra, Auli, and Zaremba]{ranzato2016exposurebias}
Marc'Aurelio Ranzato, Sumit Chopra, Michael Auli, and Wojciech Zaremba.
\newblock Sequence level training with recurrent neural networks.
\newblock In \emph{Proceedings of the 4th International Conference on Learning Representations}, 2016.

\bibitem[Rombach et~al.(2022)Rombach, Blattmann, Lorenz, Esser, and Ommer]{rombach2021stablediff}
Robin Rombach, Andreas Blattmann, Dominik Lorenz, Patrick Esser, and Bj{\"o}rn Ommer.
\newblock High-resolution image synthesis with latent diffusion models.
\newblock In \emph{Proceedings of the IEEE Conference on Computer Vision and Pattern Recognition (CVPR)}, 2022.

\bibitem[Schmidt(2019)]{schmidt2019exposurebias}
Florian Schmidt.
\newblock Generalization in generation: A closer look at exposure bias.
\newblock In \emph{Proceedings of the 3rd Workshop on Neural Generation and Translation}, page 157–167, 2019.

\bibitem[Shwartz-Ziv and Armon(2022)]{tabdl2022}
Ravid Shwartz-Ziv and Amitai Armon.
\newblock Tabular data: Deep learning is not all you need.
\newblock \emph{Information Fusion}, 81:\penalty0 84--90, 2022.

\bibitem[Sohl-Dickstein et~al.(2015)Sohl-Dickstein, Weiss, Maheswaranathan, and Ganguli]{jsd2015diffusion}
Jascha Sohl-Dickstein, Eric~A. Weiss, Niru Maheswaranathan, and Surya Ganguli.
\newblock Deep unsupervised learning using nonequilibrium thermodynamics.
\newblock In \emph{Proceedings of the 32nd International Conference on Machine Learning}. PMLR, 2015.

\bibitem[Song and Ermon(2019)]{song2019scorematching}
Yang Song and Stefano Ermon.
\newblock Generative modeling by estimating gradients of the data distribution.
\newblock In \emph{Proceedings of the 33rd Conference on Neural Information Processing Systems}, 2019.

\bibitem[Song et~al.(2021)Song, Sohl-Dickstein, Kingma, Kumar, Ermon, and Poole]{song2021scoresde}
Yang Song, Jascha Sohl-Dickstein, Diederik~P Kingma, Abhishek Kumar, Stefano Ermon, and Ben Poole.
\newblock Score-based generative modeling through stochastic differential equations.
\newblock In \emph{Proceedings of the 9th International Conference on Learning Representations}, 2021.

\bibitem[Sparapani et~al.(2021)Sparapani, Spanbauer, and McCulloch]{bartrpackage2021}
Rodney Sparapani, Charles Spanbauer, and Robert McCulloch.
\newblock Nonparametric machine learning and efficient computation with {B}ayesian {A}dditive {R}egression {T}rees: The {BART} {R} package.
\newblock \emph{Journal of Statistical Software}, 97\penalty0 (1):\penalty0 1–66, 2021.

\bibitem[Starling et~al.(2020)Starling, Murray, Carvalho, Bukowski, and Scott]{starling2020}
Jennifer~E. Starling, Jared~S. Murray, Carlos~M. Carvalho, Radek~K. Bukowski, and James~G. Scott.
\newblock {BART} with {T}argeted {S}moothing: An analysis of patient-specific stillbirth risk.
\newblock \emph{The Annals of Applied Statistics}, 14\penalty0 (1):\penalty0 28–50, 2020.

\bibitem[Ustimenko and Prokhorenkova(2021)]{sglb2021}
Aleksei Ustimenko and Liudmila Prokhorenkova.
\newblock {SGLB}: Stochastic gradient langevin boosting.
\newblock In \emph{Proceedings of the 38th International Conference on Machine Learning}. PMLR, 2021.

\bibitem[Vanschoren et~al.(2013)Vanschoren, van Rijn, Bischl, and Torgo]{OpenML2013}
Joaquin Vanschoren, Jan~N. van Rijn, Bernd Bischl, and Luis Torgo.
\newblock {OpenML}: Networked science in machine learning.
\newblock \emph{SIGKDD Explorations}, 15\penalty0 (2):\penalty0 49--60, 2013.
\newblock URL \url{http://doi.acm.org/10.1145/2641190.2641198}.

\bibitem[Verma and Nalisnick(2022)]{verma2022cl2d}
Rajeev Verma and Eric~T. Nalisnick.
\newblock Calibrated learning to defer with one-vs-all classifiers.
\newblock In \emph{Proceedings of the 39th International Conference on Machine Learning}, 2022.

\bibitem[Vincent(2011)]{vincent2011connection}
Pascal Vincent.
\newblock A connection between score matching and denoising autoencoders.
\newblock \emph{Neural Computation}, 23\penalty0 (7):\penalty0 1661--1674, 2011.

\bibitem[Wang and Zhou(2020)]{wang2020thompson}
Zhendong Wang and Mingyuan Zhou.
\newblock Thompson sampling via local uncertainty.
\newblock In \emph{Proceedings of the 37th International Conference on Machine Learning}. PMLR, 2020.

\bibitem[Watt et~al.(2020)Watt, Borhani, and Katsaggelos]{watt2020mlr}
Jeremy Watt, Reza Borhani, and Aggelos~Konstantinos Katsaggelos.
\newblock Universal approximators.
\newblock In \emph{Machine Learning Refined: Foundations, Algorithms, and Applications}, chapter 11.2. Cambridge University Press, 2020.

\bibitem[Williams and Zipser(1989)]{teacherforcing1989}
Ronald~J. Williams and David Zipser.
\newblock A learning algorithm for continually running fully recurrent neural networks.
\newblock \emph{Neural Computation}, 1\penalty0 (2):\penalty0 270--280, 1989.

\bibitem[Yang et~al.(2018)Yang, Morillo, and Hospedales]{yang2018dndt}
Yongxin Yang, Irene~Garcia Morillo, and Timothy~M. Hospedales.
\newblock Deep neural decision trees.
\newblock In \emph{2018 ICML Workshop on Human Interpretability in Machine Learning (WHI 2018)}, 2018.

\bibitem[Yeh and hui Lien(2009)]{yeh2009creditcarddefault}
I-Cheng Yeh and Che hui Lien.
\newblock The comparisons of data mining techniques for the predictive accuracy of probability of default of credit card clients.
\newblock \emph{Expert Systems with Applications}, 36\penalty0 (2, Part 1):\penalty0 2473--2480, 2009.

\bibitem[Yin and Zhou(2018)]{yin2018semi}
Mingzhang Yin and Mingyuan Zhou.
\newblock {S}emi-{I}mplicit {V}ariational {I}nference.
\newblock In \emph{Proceedings of the 35th International Conference on Machine Learning}. PMLR, 2018.

\bibitem[Yu et~al.(2023)Yu, Xie, Zhu, Yang, Zhang, and Zhang]{yu2023hierarchical}
Longlin Yu, Tianyu Xie, Yu~Zhu, Tong Yang, Xiangyu Zhang, and Cheng Zhang.
\newblock Hierarchical semi-implicit variational inference with application to diffusion model acceleration.
\newblock In \emph{Proceedings of the 37th Conference on Neural Information Processing Systems}, 2023.

\bibitem[Zhou et~al.(2024)Zhou, Zheng, Wang, Yin, and Huang]{zhou2024score}
Mingyuan Zhou, Huangjie Zheng, Zhendong Wang, Mingzhang Yin, and Hai Huang.
\newblock {S}core identity {D}istillation: Exponentially fast distillation of pretrained diffusion models for one-step generation.
\newblock In \emph{Proceedings of the 41st International Conference on Machine Learning}, 2024.

\bibitem[Zhou et~al.(2023)Zhou, Jiao, Liu, and Huang]{jointmatching}
Xingyu Zhou, Yuling Jiao, Jin Liu, and Jian Huang.
\newblock A deep generative approach to conditional sampling.
\newblock \emph{Journal of the American Statistical Association}, 118\penalty0 (543):\penalty0 1837–1848, 2023.

\end{thebibliography}
\bibliographystyle{plainnat}

\newpage
\appendix

\section{Appendix}

\subsection{Background: An In-Depth Version}
\label{ssec:background_detailed}
In this section, we provide a more comprehensive version of \cref{sec:background}, with a focus on establishing the objective functions of both gradient boosting and CARD.

\subsubsection{Supervised Learning}
\label{sssec:supervised_learning}
We aim to tackle the problem of \textit{supervised learning}: given a set of covariates $\vx=\{x_1, \dots, x_p\}$, and a response variable $\vy$ --- a numerical variable for regression, or a categorical one for classification --- we seek to learn a mapping that takes the covariates as inputs and predicts the response variable as its output, with the hope that it can generalize to new and unseen data after observing some training data.

The mapping usually takes the form of a mathematical function, thus the supervised learning problem becomes a \textit{function estimation} problem: the goal is to obtain an approximation $F^*(\vx)$ of the function $F(\vx)$ that maps $\vx$ to $\vy$, which minimizes the expectation of a loss function $L\big(\vy, F(\vx)\big)$ over the joint distribution $p(\vx, \vy)$ \citep{friedman2001gbm}: 
\ba{F^* &= \argmin_F\E_{p(\vx, \vy)}\big[L\big(\vy, F(\vx)\big)\big].}

When imposing a parametric form $\vtheta$ upon the function $F$ as a common practice, the function is now read as $F(\vx; \vtheta)$, and the function estimation problem becomes a \textit{parameter optimization} problem: \ba{\vtheta^*=\argmin_{\vtheta} \E_{p(\vx, \vy)}\big[L\big(\vy, F(\vx; \vtheta)\big)\big], \label{eq:param_optim}} thus $F^*(\vx) = F(\vx; \vtheta^*)$.

Numerical optimization methods need to be applied to solve Eq.\,(\ref{eq:param_optim}) for most $F(\vx; \vtheta)$ and $L$ \citep{friedman2001gbm}. The standard procedure for many of these methods is as follows: first, they determine the direction to improve the objective $L$, then compute the step size via line search \citep{nocedalwright1999numopt} for the parameter $\vtheta$ to move along this direction. Gradient descent \citep{cauchy1847} is one of the most well-known methods in the machine learning community for finding the descent direction, which computes \textit{the negative gradient} as the steepest descent direction for an objective function differentiable in the neighborhood of the point of interest.

\subsubsection{Gradient Descent}
\label{sssec:gradient_descent}
Denote the objective function in Eq.\,(\ref{eq:param_optim}) as $$\Phi(\vtheta)=\E_{p(\vx, \vy)}\big[L\big(\vy, F(\vx; \vtheta)\big)\big],$$ the gradient descent update at any intermediate step $m$ is \ba{\vtheta_m = \vtheta_{m-1}+\rho_m\cdot\big(-\nabla_{\vtheta_{m-1}}\Phi(\vtheta_{m-1})\big), \label{eq:gd_update}} where the step size \ba{\rho_m=\argmin_{\rho}\Phi\Big(\vtheta_{m-1}+\rho\cdot\big(-\nabla_{\vtheta_{m-1}}\Phi(\vtheta_{m-1})\big)\Big).}

Therefore, with $M$ total update steps and $\vtheta_0$ as the initialization, we have the optimized parameter via gradient descent as \ba{\vtheta^*=\vtheta_0 + \sum_{m=1}^M\rho_m\cdot\big(-\nabla_{\vtheta_{m-1}}\Phi(\vtheta_{m-1})\big). \label{eq:optim_param_gd}}

\subsubsection{Gradient Boosting}
\label{sssec:gradient_boosting}
While gradient descent can be described as a numerical optimization method \textit{in the parameter space}, gradient boosting \citep{friedman2001gbm} is essentially gradient descent \textit{in the function space}: by considering $F(\vx)$ evaluated at each $\vx$ to be a parameter, \citet{friedman2001gbm} establishes the objective function at the joint distribution level as 
\ba{\Phi(F) = \E_{p(\vx, \vy)}\big[L\big(\vy, F(\vx)\big)\big] = \E_{p(\vx)}\bigg[\E_{p(\vy\given\vx)}\big[L\big(\vy, F(\vx)\big)\big]\bigg], \label{eq:function_expected_loss}}
and equivalently, the objective function at the instance level: \ba{\Phi\big(F(\vx)\big)=\E_{p(\vy\given\vx)}\big[L\big(\vy, F(\vx)\big)\big],} whose gradient can be computed as 
\ba{\nabla_{F(\vx)}\Phi\big(F(\vx)\big)=\frac{\partial \Phi\big(F(\vx)\big)}{\partial F(\vx)} =\E_{p(\vy\given\vx)}\left[\frac{\partial L\big(\vy, F(\vx)\big)}{\partial F(\vx)}\right], \label{eq:instance_function_gradient}}
where the second equation results from assuming sufficient regularity to interchange differentiation and integration.

Following the gradient-based numerical optimization paradigm as in Eq.\,(\ref{eq:optim_param_gd}), we obtain the optimal solution in the function space: \ba{F^*(\vx)=f_0(\vx) + \sum_{m=1}^M\rho_m\cdot\big(-g_m(\vx)\big), \label{eq:optim_func_gd}} where $f_0(\vx)$ is the initial guess, and $g_m(\vx)=\nabla_{F_{m-1}(\vx)}\Phi\big(F_{m-1}(\vx)\big)$ is the gradient at optimization step $m$.

Given a finite set of samples $\{\vy_i, \vx_i\}_1^N$ from $p(\vx, \vy)$, we have the data-based analogue of $g_m(\vx)$ defined only at these training instances: \ba{g_m(\vx_i)=\frac{\partial L\big(\vy_i, \hat{F}_{m-1}(\vx_i)\big)}{\partial \hat{F}_{m-1}(\vx_i)}.} Since the goal of supervised learning is to generalize the predictive function to unseen data, \citet{friedman2001gbm} proposes to use a parameterized class of functions $h(\vx; \valpha)$ to learn the negative gradient term at every gradient descent step. Specifically, $h(\vx; \valpha)$ is trained with the squared-error loss at optimization step $m$ to produce $\{h(\vx_i; \valpha_m)\}_1^N$ most parallel to $\{-g_m(\vx_i)\}_1^N$, and the solution $h(\vx;\valpha_m)$ can be applied to approximate $-g_m(\vx)$ for any $\vx$, whose parameter \ba{\valpha_m=\argmin_{\vxi, \omega}\sum_{i=1}^N\big(-g_m(\vx_i)-\omega\cdot h(\vx_i;\vxi)\big)^2, \label{eq:grad_pred}} while the multiplier $\rho_m$ is optimized via line search, 
\ba{\rho_m=\argmin_{\rho}\sum_{i=1}^N L\big(\vy_i, \hat{F}_{m-1}(\vx_i)+\rho\cdot h(\vx_i;\valpha_m)\big).}

Therefore, with finite data, the gradient descent update in the function space at step $m$ is \ba{\hat{F}_m(\vx) = \hat{F}_{m-1}(\vx)+\rho_m\cdot h(\vx;\valpha_m), \label{eq:gb_one_step}} and the prediction of $\vy$ given any $\vx$ can be obtained through \ba{\hat{\vy}=\hat{F}^*(\vx) = \hat{F}_0(\vx)+\sum_{m=1}^M\rho_m\cdot h(\vx;\valpha_m). \label{eq:pred_gb}}

The function $h(\vx; \valpha)$ is termed a \textit{weak learner} or \textit{base learner}, and is often parameterized by a simple Classification And Regression Tree (CART) \citep{breiman84cart}. Eq.\,(\ref{eq:pred_gb}) has the form of an ensemble of weak learners, trained sequentially and combined via weighted sum\footnote{Each weight $\rho_m$ is conceptually the step size in numerical optimization. In practice, we often find it to be a constant that is preset or scheduled, instead of learned through line search: \textit{e.g.}, in \citet{ke2017lightgbm}, the weight of each weak learner is set to $1$.}.

Among all choices of the loss function $L\big(\vy, F(\vx)\big)$, the squared-error loss is of particular interest: \ba{L\big(\vy, F(\vx)\big)=\tfrac{1}{2}\big(\vy - F(\vx)\big)^2. \label{eq:sq_loss}} In this case, the negative gradient is \ba{-\frac{\partial L}{\partial F(\vx)}=\vy-F(\vx),} which is the residual\footnote{We attach the algorithm of gradient boosting with the squared-error loss in \cref{ssec:gb_ls_reg} for reference.}. As a result, each weak learner aims to predict the residual term at its corresponding optimization step. It is tempting to draw parallels between this residual predicting behavior by gradient boosting and the residual learning paradigm by ResNet \citep{he2015resnet} at face value, thus we intend to point out their difference here: the former is due to the particular choice of the squared-error loss function, Eq.\,(\ref{eq:sq_loss}), while the latter results from its computational excellencies in dealing with the vanishing/exploding gradient problem in very deep neural networks.

The squared-error loss is the default choice of loss function for regression tasks by popular gradient boosting libraries like XGBoost \citep{chen2016xgboost}, LightGBM \citep{ke2017lightgbm}, and CatBoost \citep{lp2018catboost}. It is worth mentioning that the optimal solution for minimizing the expected square-error loss is the conditional mean, $\E[\vy\given\vx]$.

\subsubsection{Classification and Regression Diffusion Models (CARD)}
\label{sssec:CARD}
With the same goal as gradient boosting of taking on supervised learning problems, CARD \citep{han2022card} approaches them from a very different angle: by adopting a generative modeling framework, a CARD model directly outputs the samples from the conditional distribution $p(\vy\given\vx)$, instead of some summary statistics such as the conditional mean $\E[\vy\given\vx]$. A unique advantage of this class of models is that it is free of any assumptions on the parametric form of $p(\vy\given\vx)$ --- \textit{e.g.}, the additive-noise assumption with a particular form of its noise distribution (a zero-mean Gaussian distribution for regression, or a standard Gumbel distribution for classification), which has been prevalently applied by the existing methods. A finer level of granularity in the outputs of CARD (\textit{i.e.}, directly generating samples instead of predicting a summary statistic) helps to paint a more complete picture of $p(\vy\given\vx)$: with enough samples, the model can capture the variability and modality of the conditional distribution, besides accurately recovering $\E[\vy\given\vx]$. The advantage of generative modeling becomes more evident when $p(\vy\given\vx)$ is multimodal, or has heteroscedasticity, as shown by the toy examples in \citet{han2022card}. Meanwhile, CARD consistently performs better in terms of the conventional metrics like RMSE and NLL on real-world datasets than other uncertainty-aware methods that are explicitly optimized for these metrics as their objectives.

The parameterization of CARD follows the Denoising Diffusion Probabilistic Models (DDPM) \citep{ho2020ddpm} framework, which is a generative model that aims to learn a function that maps a sample from a simple known distribution (often called the \textit{noise distribution}) to a sample from the target distribution. However, instead of directly generating the sample with only one function evaluation like other classes of generative models --- including GANs \citep{goodfellow2014gan} and VAEs \citep{kingma2014vae} --- the function produces a \textit{less noisy version} of its input after each evaluation, which is then fed into the \textit{same} function to produce the next one. For CARD, the final output can be viewed as a noiseless sample of $\vy$ from $p(\vy\given\vx)$ after enough steps. This autoregressive fashion of computing can be described as \textit{iterative refinement} or \textit{progressive denoising}.

CARD adopts the DDPM framework by treating the noisy samples from the intermediate steps as latent variables, and construct a Markov chain to link them together, so that the progressive data \textit{generation} process can be modeled analytically, in the sense that an explicit distributional form (\textit{i.e.}, Gaussian) can be imposed upon adjacent latent variables.

This Markov chain is formed in the direction \textit{opposite} to the data \textit{generation} process described above, with each variable subscripted by its chronological order: \textit{e.g.}, the target response variable $\vy$ is re-denoted as $\vy_0$, and the noise variable is $\vy_T$, where $T$ is the total number of steps, or timesteps, for this Markov process. As the data generation process that goes from $\vy_T$ to $\vy_0$ is described as a denoising procedure above, this Markov chain that goes from $\vy_0$ to $\vy_T$ defines a noise-adding mechanism, where the stepwise transition $q(\vy_t\given\vy_{t-1}, \vx)$ is defined through a Gaussian distribution. The conditional distribution of all latent variables given the target variable (and the covariates) in the noise-adding direction, \ba{q(\vy_{1:T}\given\vy_0, \vx)=\prod_{t=1}^T q(\vy_t\given\vy_{t-1}, \vx),} is called the \textit{forward diffusion process}.

Meanwhile, denoting the learnable parameter in the generative model as $\vtheta$, the joint distribution (conditioning on the covariates) in the data generation direction is \ba{p_{\vtheta}(\vy_{0:T}\given\vx)=p(\vy_T\given\vx)\prod_{t=1}^T p_{\vtheta}(\vy_{t-1}\given\vy_t, \vx),} in which $p(\vy_T\given\vx)=\gN(\vmu_T, \mI)$ is \textit{the} noise distribution --- a Gaussian distribution with a mean of $\vmu_T$ --- and is also referred to as the \textit{prior distribution}. This joint distribution is called the \textit{reverse diffusion process}.

As a generative model, CARD is trained via an objective rooted in distribution matching: re-denoting the ground truth conditional distribution $p(\vy\given\vx)$ as $q(\vy_0\given\vx)$, we wish to learn $\vtheta$ so that $p_{\vtheta}(\vy_0\given\vx)$ approximates $q(\vy_0\given\vx)$ well, \textit{i.e.}, \ba{\kldiv[\big]{q(\vy_0\given\vx)}{p_{\vtheta}(\vy_0\given\vx)}\approx 0, \label{eq:kl_obj}} where \ba{p_{\vtheta}(\vy_0\given\vx)=\int p_{\vtheta}(\vy_{0:T}\given\vx)d\vy_{1:T}.} We have the following relationship: \ba{H(q, p_{\vtheta}) = H(q)+\kldiv[]{q}{p_{\vtheta}},} in which $H(q, p_{\vtheta})=-\E_q[\log p_{\vtheta}]$ is the cross entropy of $p_{\vtheta}(\vy_0\given\vx)$, $H(q)=\E_q[-\log q]$ is the entropy of $q(\vy_0\given\vx)$, and $\kldiv[]{q}{p_{\vtheta}}=\E_q[\log q/p_{\vtheta}]$ is the KL divergence of $q(\vy_0\given\vx)$ from $p_{\vtheta}(\vy_0\given\vx)$. Since $q(\vy_0\given\vx)$ does not contain $\vtheta$, $\min_{\vtheta}\kldiv[\big]{q(\vy_0\given\vx)}{p_{\vtheta}(\vy_0\given\vx)}$ is equivalent to $\min_{\vtheta}H\big(q(\vy_0\given\vx), p_{\vtheta}(\vy_0\given\vx)\big)$. The variational bound (\textit{i.e.}, the negative ELBO) can be derived from this cross entropy term (see Appendix \ref{ssec:card_neg_elbo}) as a standard objective function to be minimized for training CARD: \ba{&\E_{q(\vy_0\given\vx)}\big[-\log p_{\vtheta}(\vy_0\given\vx)\big] \notag \\ \leq &\E_{q(\vy_{0:T}\given\vx)}\left[\log\frac{q(\vy_{1:T}\given\vy_0, \vx)}{p_{\vtheta}(\vy_{0:T}\given\vx)}\right]\eqqcolon L \label{eq:card_obj}.} By following the same procedure as Appendix A in \citet{ho2020ddpm}, the objective $L$ in (\ref{eq:card_obj}) can be rewritten as \ba{L=\E_{q(\vy_{0:T}\given\vx)}\left[L_T+\sum_{t=2}^TL_{t-1}+L_0\right], \label{eq:obj_rewrite}} in which \bastar{&L_T \coloneqq \kldiv[\big]{q(\vy_T\given\vy_0,\vx)}{p(\vy_T\given\vx)}, \\ &L_{t-1} \coloneqq \kldiv[\big]{q(\vy_{t-1}\given\vy_t,\vy_0,\vx)}{p_{\vtheta}(\vy_{t-1}\given\vy_t,\vx)}, \\ &L_0 \coloneqq -\log p_{\vtheta}(\vy_0\given\vy_1,\vx).}

As what will be shown later, the forward process does not contain any learnable parameters, thus $L_T$ is a constant with respect to $\vtheta$. Meanwhile, the form of $p_{\vtheta}(\vy_0\given\vy_1,\vx)$ is more of an application-dependent design choice. Therefore, the main focus for optimizing $\vtheta$ is on the remaining $L_{t-1}$ terms, for $t=2,\dots,T$.

The distribution $q(\vy_{t-1}\given\vy_t,\vy_0,\vx)$ in $L_{t-1}$ is called the \textit{forward process posterior distribution}, which is tractable and can be derived by applying Bayes' rule: 
\ba{q(\vy_{t-1}\given\vy_t, \vy_0, \vx) \propto q\big(\vy_t\given\vy_{t-1}, \vx\big)\cdot q\big(\vy_{t-1}\given\vy_0, \vx\big),} 
in which both $q\big(\vy_t\given\vy_{t-1}, \vx\big)$ and $q\big(\vy_{t-1}\given\vy_0, \vx\big)$ are Gaussian: as mentioned before, the former is the stepwise transition distribution in the forward process, defined as 
\ba{q(\vy_t\given\vy_{t-1}, \vx) =\gN\big(\vy_t; \sqrt{\alpha_t}\vy_{t-1} + (1-\sqrt{\alpha_t})\vmu_T, \beta_t\mI\big),}
in which $\beta_t$ is the $t$-th term of a predefined noise schedule $\beta_1,\dots,\beta_T$, and $\alpha_t \coloneqq 1-\beta_t$. This design gives rise to a closed-form distribution to sample $\vy_t$ at any arbitrary timestep $t$: 
\ba{q(\vy_t\given\vy_0, \vx) =\gN\big(\vy_t; \sqrt{\bar{\alpha}_t}\vy_0 + (1-\sqrt{\bar{\alpha}_t})\vmu_T, (1-\bar{\alpha}_t)\mI\big), \label{eq:card_arb_fwd}}
in which $\bar{\alpha}_t \coloneqq \prod_{j=1}^t\alpha_j$. Each of the forward process posteriors thus has the form of 
\ba{q(\vy_{t-1}\given\vy_t, \vy_0, \vx)=\gN\Big(\vy_{t-1}; \tilde{\vmu}(\vy_t, \vy_0, \vmu_T), \tilde{\beta_t}\mI\Big), \label{eq:card_post}}
where the variance \ba{\tilde{\beta_t} \coloneqq \frac{1-\bar{\alpha}_{t-1}}{1-\bar{\alpha}_t}\beta_t, \label{eq:card_post_var}} and the mean \ba{\tilde{\vmu}(\vy_t, \vy_0, \vmu_T) \coloneqq \gamma_0\cdot\vy_0+\gamma_1\cdot\vy_t+\gamma_2\cdot\vmu_T, \label{eq:card_post_mean}} in which the coefficients are: \bastar{&\gamma_0=\frac{\beta_t\sqrt{\bar{\alpha}_{t-1}}}{1-\bar{\alpha}_t}, \\ &\gamma_1=\frac{(1-\bar{\alpha}_{t-1})\sqrt{\alpha_t}}{1-\bar{\alpha}_t}, \\ &\gamma_2=\Bigg(1+\frac{(\sqrt{\bar{\alpha}_t}-1)(\sqrt{\alpha_t}+\sqrt{\bar{\alpha}_{t-1}})}{1-\bar{\alpha}_t}\Bigg),} as derived in Appendix A.1 of \citet{han2022card}.

Now to minimize each $L_{t-1}$, $p_{\vtheta}(\vy_{t-1}\given\vy_t,\vx)$ needs to approximate the Gaussian distribution $q(\vy_{t-1}\given\vy_t,\vy_0,\vx)$, whose variance (\ref{eq:card_post_var}) is already known. Therefore, the learning task is reduced to optimizing $\vtheta$ for the estimation of the forward process posterior mean $\tilde{\vmu}(\vy_t, \vy_0, \vmu_T)$. CARD adopts the noise-prediction loss introduced in \citet{ho2020ddpm}, a simplification of $L_{t-1}$: 
\ba{\gL_{\text{CARD}}=\E_{p(t,\vy_0\given\vx,\epsilonv)}\left[\big|\big|\epsilonv - \epsilonv_{\vtheta}\big(\vx, \vy_t, f_{\phi}(\vx), t\big)\big|\big|^2\right], \label{eq:noise_est}}
in which $\epsilonv\sim\gN(\vzero, \mI)$ is sampled as the \textit{forward} process noise term, $\vy_t=\sqrt{\bar{\alpha}_t}\vy_0+(1-\sqrt{\bar{\alpha}_t})\vmu_T+\sqrt{1-\bar{\alpha}_t}\epsilonv$ is the sample from the \textit{forward} process distribution (\ref{eq:card_arb_fwd}), and $f_{\phi}(\vx)$ is the point estimate of $\E[\vy\given\vx]$ (usually parameterized by a pre-trained neural network, and to be used as an additional input to the diffusion model $\epsilonv_{\vtheta}$).

Note that the vanilla CARD framework directly sets the prior mean as $\vmu_T=f_{\phi}(\vx)$. Here we write the generic form $\vmu_T$ instead, not only for better clarity in methodology demonstration, but also as a slight design change in the diffusion boosting framework: we introduce an extra degree of freedom into the vanilla CARD framework by allowing the choice of the prior mean to be different than the conditional mean estimation $f_{\phi}(\vx)$, while still using $f_{\phi}(\vx)$ as one input to the diffusion model at each timestep since it possesses the information of $\E[\vy\given\vx]$.

\subsection{In-Depth Analysis of Related Studies}
\label{ssec:related_work_in_depth}
In this section, we contextualize our work by exploring its relationships with several related studies referenced in \cref{sec:related_work}.

\subsubsection{Distinctions Between Diffusion Boosting and SGLB}
\label{sssec:db_vs_sglb}
\citet{sglb2021} introduces SGLB, a gradient boosting algorithm based on the Langevin diffusion equation. Despite the similar terminology used for their names, SGLB and Diffusion Boosting fundamentally differ in their methodologies. To clarify these differences, we have summarized the key distinctions between SGLB and Diffusion Boosting in \cref{tab:db_vs_sglb}.

\begin{table}[H]
\caption{\label{tab:db_vs_sglb}Differences between SGLB and Diffusion Boosting.}\vspace{1mm}
\begin{center}
\resizebox{16cm}{!}{
\begin{tabular}{@{}c|c|c@{}}
\toprule[1.05pt] \hline
 & SGLB & Diffusion Boosting \\ \hline\hline %
Is the method a variant of gradient boosting? & yes & no \\ \hline
target of the weak learner $h_t$ for different $t$'s & different & same \\ \hline
input of the weak learner $h_t$ for different $t$'s & same & different \\ \hline
objective function for regression and for classification & different & same \\ \hline
presence of stochasticity & only during training & during both training and inference \\ \hline
Is the output of the trained model deterministic? & yes & no \\ \hline
output of the model & a point estimate of $\E[\vy\given\vx]$ or $p(\vy=1\given\vx)$ & a sample from $p(\vy\given\vx)$ \\ \hline
context of the term ``diffusion'' & a special form of the Langevin diffusion equation & diffusion models as a class of generative models \\ \hline
\bottomrule[1.05pt]
\end{tabular}
}
\end{center}
\end{table}

We elaborate the differences listed in \cref{tab:db_vs_sglb} as follows:

\begin{itemize}[noitemsep, topsep=0pt, leftmargin=*]
    \item SGLB is a variant of gradient boosting, while Diffusion Boosting is not: 
    \begin{itemize}
        \item SGLB builds upon the original gradient boosting framework by adding a Gaussian noise sample (whose variance is controlled by the inverse diffusion temperature hyperparameter) to the negative gradient to form the target of each weak learner, which helps the numerical optimization to achieve convergence to the global optimum, regardless of the convexity of the loss function. In other words, SGLB is a gradient boosting algorithm that achieves better global convergence than the original algorithm. In the vanilla gradient boosting, the objective function of the weak learner $h_t(\vx)$ at step $t$ is $$||-\nabla_{F_{(t-1)}(\vx)} L\big(\vy, F_{(t-1)}(\vx)\big)-h_t(\vx)||^2,$$ \textit{i.e.}, the weak learners across different $t$'s have \textit{different} targets $-\nabla_{F_{(t-1)}(\vx)} L\big(\vy, F_{(t-1)}(\vx)\big)$ to predict, but share the \textit{same} input $\vx$.
        \item In Diffusion Boosting, the objective function of the weak learner $h_t\big(\hat{\vy}_t, \vx, f_{\phi}(\vx)\big)$ is $$|| \vy-h_t\big(\hat{\vy}_t, \vx, f_{\phi}(\vx)\big)||^2,$$ \textit{i.e.}, the weak learners across different steps share the \textit{same} target $\vy$ to predict, but have \textit{different} inputs $\hat{\vy}_t$ (See \cref{alg:dbt_train} Line 11).
    \end{itemize}
    \item Objective function: 
    \begin{itemize}
        \item SGLB uses different objective functions $L\big(\vy, F(\vx)\big)$ for regression and for classification: $L$ is usually chosen to be the squared-error loss for regression, and logistic loss for classification.
        \item Diffusion Boosting uses the same objective function for both types of supervised learning tasks: \newline$\kldiv[\big]{q(\vy\given\vx)}{p_{\vtheta}(\vy\given\vx)}$ as in Eq.\,(\ref{eq:kl_obj}), or equivalently, $|| \vy-h_t\big(\hat{\vy}_t, \vx, f_{\phi}(\vx)\big)||^2$ for each weak learner.
    \end{itemize}
    \item Stochasticity is only present during the training of SGLB, but is present during both training and inference of Diffusion Boosting: 
    \begin{itemize}
        \item For SGLB, stochasticity is introduced during training in the form of a Gaussian noise sample added to each negative gradient, in order to facilitate parameter space exploration. Once the model is trained, the prediction is the same given the same covariates $x$: a \textit{point estimate} of $\E[\vy\given\vx]$ for regression when the objective is the squared-error loss, or of $p(\vy=1 \given \vx)$ for classification when the objective is the logistic loss.
        \item For Diffusion Boosting, stochasticity is present during training when sampling $\vy_{t+1}$ via the forward process (\cref{alg:dbt_train} Line 6) and $\hat{\vy}_t$ via the posterior (\cref{alg:dbt_train} Line 9), as well as inference (\cref{alg:dbt_inference} Line 1 and Line 5). Given the same covariates $\vx$, the output is different across different draws, each represents a sample from the learned $p(\vy \given \vx)$.
    \end{itemize}
    \item The context of ``diffusion'':
    \begin{itemize}
        \item In SGLB, the word ``diffusion'' appeared via terms ``Langevin diffusion'' and ``inverse diffusion temperature'', both of which are related to the mathematical description of the evolution of particles over time, under the context of physics or stochastic processes.
        \item In Diffusion Boosting, the word ``diffusion'' is short for ``diffusion models'' as a class of generative models proposed by \citet{jsd2015diffusion}.
        \item Note that \citet{song2021scoresde} provides an alternative formulation of diffusion models via stochastic SDE, whose forward SDE Eq.\,(5) resembles the one shown in \citet{sglb2021} Eq.\,(6), but the data generation process refers to the reverse SDE, \textit{i.e.}, Eq.\,(6) in \citet{song2021scoresde}.
    \end{itemize}
\end{itemize}

\subsubsection{Distinctions Between Diffusion Boosting and GBDT Ensembles}
\label{sssec:db_vs_gbdt_ensembles}
\citet{gbdtuncertainty2021} propose GBDT ensembles, an ensemble-based framework parameterized by trees, designed to estimate predictive uncertainty in supervised learning tasks, specifically targeting epistemic uncertainty. We have summarized the key differences between GBDT ensembles and Diffusion Boosting in \cref{tab:db_vs_gbdt_ensembles}.

\begin{table}[H]
\caption{\label{tab:db_vs_gbdt_ensembles}Differences between GBDT ensembles and Diffusion Boosting.}\vspace{1mm}
\begin{center}
\resizebox{16cm}{!}{
\begin{tabular}{@{}c|c|c@{}}
\toprule[1.05pt] \hline
 & GBDT ensembles & Diffusion Boosting \\ \hline\hline %
Parametric assumption on $p(\vy \vert \vx)$? & yes & no \\ \hline
types of predictive uncertainty estimated & total uncertainty, including both aleatoric and epistemic uncertainty & only aleatoric uncertainty \\ \hline
OOD detection & yes & no \\ \hline
\bottomrule[1.05pt]
\end{tabular}
}
\end{center}
\end{table}

We elaborate the differences listed in \cref{tab:db_vs_gbdt_ensembles} as follows:

\begin{itemize}[noitemsep, topsep=0pt, leftmargin=*]
    \item Assumption on the parametric form of the distribution $p(\vy\given\vx)$: 
    \begin{itemize}
        \item To achieve uncertainty estimation in regression, GBDT ensembles assumes $p(\vy\given\vx)$ to have a Gaussian distribution, whose parameters (mean and log standard deviation) are optimized via the expected Gaussian NLL.
        \item Diffusion Boosting does not assume any parametric form on $p(\vy\given\vx)$. This is a nontrivial paradigm shift from most existing supervised learning methods, which provides additional versatility in modeling conditional distributions, including the ones with multimodality and heteroscedasticity, as shown in our toy regression examples (\cref{fig:dbm_vs_card_toy}).
    \end{itemize}
    \item Types of predictive uncertainty each method focuses on modeling:
    \begin{itemize}
        \item GBDT ensembles model both aleatoric uncertainty (data uncertainty) and epistemic uncertainty (knowledge uncertainty) via the decomposition of uncertainty \citep{uncertaintydecomp2018} for both classification and regression, with an emphasis on epistemic uncertainty since the experiments focus on OOD and error detection.
        \item Diffusion Boosting follows the paradigm in CARD and only models the aleatoric uncertainty, \textit{i.e.}, recovering the uncertainty inherent to the ground truth data generation mechanism. (We did experiment with OOD data: for toy examples with 1D $x$ variable, we could only observe variations at the boundary of $x$, but the model outputs a constant value outside the boundary. This aligns with the description in \citet{gbdtuncertainty2021}: ``$\dots$ as decision trees are discriminative functions, if features have values outside the training domain, then the prediction is the same as for the `closest' elements in the dataset. In other words, the models' behavior on the boundary of the dataset is further extended to the outer regions.'')
    \end{itemize}
\end{itemize}

\subsection{Gradient Boosting for Least-Squares Regression}
\label{ssec:gb_ls_reg}
We include the algorithm from \citep{friedman2001gbm} here as \cref{{alg:gb_sq_loss}} for reference, with a slight adjustment in notation.

\begin{algorithm}[h]
\begin{algorithmic}[1]
\small
\REQUIRE Training samples $\{\vy_i, \vx_i\}_1^N$
\ENSURE Trained weak learners $\{h(\vx;\valpha_m)\}_1^M$
\STATE Initialize the function by $\hat{F}_0(\vx)=\argmin_s\sum_{i=1}^N L(\vy_i, s)=\tfrac{1}{N}\sum_{i=1}^N \vy_i$
\FOR{$m=1$ to $M$}
\STATE Compute the negative gradient (a.k.a. pseudoresponses) $$\tilde{\vy}_i=-\left[\frac{\partial L\big(\vy_i, F(\vx_i)\big)}{\partial F(\vx_i)}\right]_{F(\vx)=\hat{F}_{m-1}(\vx)}=\vy_i-\hat{F}_{m-1}(\vx_i), \qquad i=1,\dots,N$$
\STATE $(\rho_m, \valpha_m)=\argmin_{\valpha, \rho}\sum_{i=1}^N\big(\tilde{\vy}_i-\rho\cdot h(\vx_i;\valpha)\big)^2$
\STATE $\hat{F}_m(\vx_i)=\hat{F}_{m-1}(\vx_i)+\rho_m\cdot h(\vx_i;\valpha_m), \qquad i=1,\dots,N$
\ENDFOR
\end{algorithmic}
\caption{Gradient Boosting on Squared-Error Loss}
\label{alg:gb_sq_loss}
\end{algorithm}

\subsection{Derivation of the Variational Bound as the Objective to Train CARD Models}
\label{ssec:card_neg_elbo}
\ba{
H\big(q(\vy_0\given\vx), p_{\vtheta}(\vy_0\given\vx)\big) &= -\E_{q(\vy_0\given\vx)}\big[\log p_{\vtheta}(\vy_0\given\vx)\big] \\
&= -\E_{q(\vy_0\given\vx)}\Big[\log\big(\int p_{\vtheta}(\vy_{0:T}\given\vx)d\vy_{1:T}\big)\Big] \\
&= -\E_{q(\vy_0\given\vx)}\Big[\log\big(\int q(\vy_{1:T}\given\vy_0, \vx)\cdot\frac{p_{\vtheta}(\vy_{0:T}\given\vx)}{q(\vy_{1:T}\given\vy_0, \vx)}d\vy_{1:T}\big)\Big] \\
&= -\E_{q(\vy_0\given\vx)}\left[\log\left(\E_{q(\vy_{1:T}\given\vy_0, \vx)}\bigg[\frac{p_{\vtheta}(\vy_{0:T}\given\vx)}{q(\vy_{1:T}\given\vy_0, \vx)}\bigg]\right)\right] \label{eq:jensenineqpart1}\\
&\leq \underbrace{-\E_{q(\vy_{0:T}\given\vx)}\left[\log\frac{p_{\vtheta}(\vy_{0:T}\given\vx)}{q(\vy_{1:T}\given\vy_0, \vx)}\right]}_{\text{negative ELBO}}, \label{eq:jensenineqpart2}
} in which we apply Jensen's inequality to go from (\ref{eq:jensenineqpart1}) to (\ref{eq:jensenineqpart2}).

\subsection{Training and Sampling Algorithms of CARD-T}
\label{ssec:card_t_algo}
We present the training and sampling algorithms of CARD-T --- which can be viewed as the non-amortized tree-based version of CARD --- in Algorithms \ref{alg:card_t_train} and \ref{alg:card_t_inference}, respectively, for reference. Note that the training of each $f_{\vtheta_{t}}$ in \cref{alg:card_t_train} can be parallelized conceptually; the for loop is included for simplicity and clarity.

\begin{algorithm}[H]
\begin{algorithmic}[1]
\small  %
\REQUIRE Training set $\{(\vx_i, \vy_{0,i})\}_{i=1}^N$
\ENSURE Trained mean estimator $f_{\phi}(\vx)$ and tree ensemble $\{f_{\vtheta_{t}}\}_{t=1}^T$
\STATE Pre-train $f_{\phi}(\vx)$ to estimate $\E[\vy_0\given\vx]$
\FOR{$t=T$ to $1$}
    \STATE Sample $\textcolor{red}{\epsilonv_t}\sim\gN(\bm{0}, \mI)$
    \STATE Obtain $\textcolor{blue}{\vy_t}$ sample via $q(\vy_t\given\vy_0, \vx)$ reparameterization: \bastar{\textcolor{blue}{\vy_t}=\sqrt{\bar{\alpha}_t}\vy_0 + (1-\sqrt{\bar{\alpha}_t})\vmu_T+\sqrt{1-\bar{\alpha}_t}\textcolor{red}{\epsilonv_t}}
\STATE Train $f_{\vtheta_{t}}$ with MSE loss to predict the forward process noise sample: \bastar{\mathcal{L}_{\vtheta}^{(t)} = \E\left[\big|\big|\textcolor{red}{\epsilonv_t} - f_{\vtheta_{t}}\big(\textcolor{blue}{\vy_t}, \vx, f_{\phi}(\vx)\big)\big|\big|^2\right]}
\ENDFOR
\end{algorithmic}
\caption{CARD-T Training}
\label{alg:card_t_train}
\end{algorithm}
\begin{algorithm}[H]
\begin{algorithmic}[1]
\small  %
\REQUIRE Test data $\{\vx_j\}_{j=1}^M$, trained $f_{\phi}(\vx)$ and $\{f_{\vtheta_{t}}\}_{t=1}^T$
\ENSURE Response variable prediction $\hat{\vy}_{0,1}$
\STATE Draw $\hat{\vy}_T\sim\mathcal{N}(\vmu_T, \mI)$
\FOR{$t=T$ to $1$}
\STATE Predict the forward process noise term $\hat{\epsilonv}_{t}=f_{\vtheta_{t}}\big(\hat{\vy}_t, \vx, f_{\phi}(\vx)\big)$
\STATE Compute $\hat{\vy}_{0,t}$ via $q(\vy_t\given\vy_0, \vx)$ reparameterization: $$\hat{\vy}_{0,t}=\frac{1}{\sqrt{\bar{\alpha}_t}}\Big(\hat{\vy}_t-(1-\sqrt{\bar{\alpha}_t})\vmu_T-\sqrt{1-\bar{\alpha}_t}\hat{\epsilonv}_{t}\Big)$$
\IF {$t > 1$}
    \STATE Draw the noisy sample $\hat{\vy}_{t-1}\sim q\big(\vy_{t-1}\given\hat{\vy}_t, \hat{\vy}_{0,t}, f_{\phi}(\vx)\big)$
\ENDIF
\ENDFOR
\STATE \textbf{return} $\hat{\vy}_{0,1}$
\end{algorithmic}
\caption{CARD-T Sampling}
\label{alg:card_t_inference}
\end{algorithm}

\subsection{Feature Importance Analysis on OpenML Dataset}
\label{ssec:feature_importance}
\begin{figure}[h]
\begin{center}
\centerline{\includegraphics[width=1.0\textwidth]{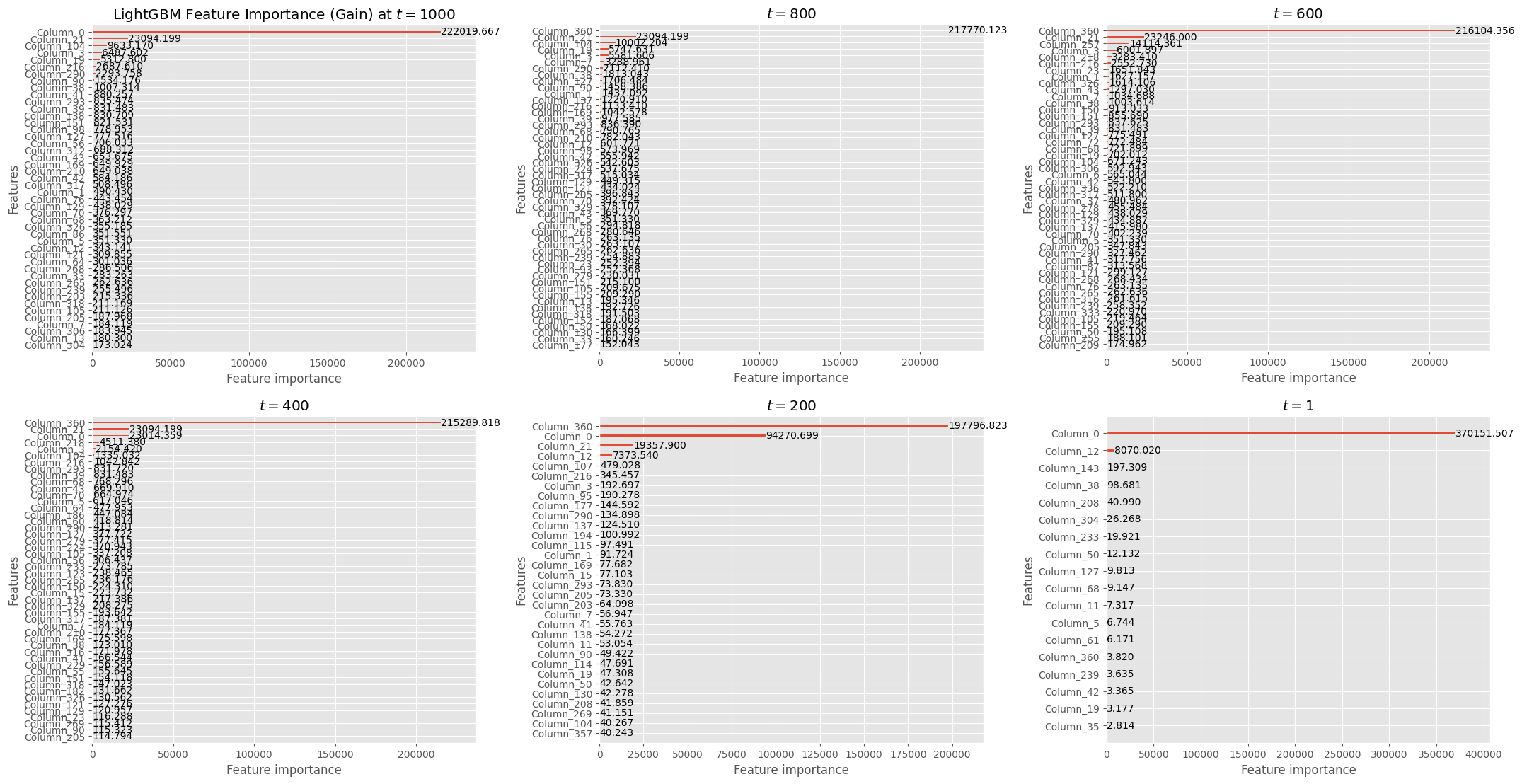}}
\caption{Feature importance plots at six diffusion timesteps.}
\label{fig:openml_feature_importance}
\end{center}
\end{figure}

We produced feature importance plots at the same timesteps as in \cref{fig:openml_shap} for the trained DBT models on the \textit{Mercedes} dataset (Table~\ref{tab:reg_openml_metrics}), as shown in \cref{fig:openml_feature_importance}. In each plot, the features are sorted by their magnitude of feature importance. We observe that the most impactful feature in Figure \ref{fig:openml_shap} and \ref{fig:openml_feature_importance} matches up at all selected timesteps. However, the remaining lists of influential features differ slightly. This discrepancy arises from the different methods used to measure feature impact:

\begin{itemize}[noitemsep, topsep=0pt, leftmargin=*]
\item \textbf{SHAP values}: SHAP values indicate how much each feature contributes to the deviation of the prediction from the average prediction (baseline). In other words, SHAP values represent a feature's responsibility for a change in the model output.
\item \textbf{Feature importance (gain)}: Feature importance based on ``gain'' measures the total improvement in the model's performance brought by a feature across all splits where the feature is used, where ``gain'' represents the reduction in the loss function when the feature is used to split the data. In other words, gain-based feature importance sums up the improvements in the loss function due to splits involving the feature.
\end{itemize}

In summary, SHAP values and feature importance provide two distinct ways of measuring the impact of features: SHAP values focus on changes in the model output, while feature importance based on gain considers improvements in the objective function.

\subsection{UCI Regression Experiment Setup}
\label{ssec:uci_reg_experiment_setup}
The same $10$ UCI regression benchmark datasets and the experimental protocol proposed in \citet{pbp}, and followed by \citet{mcdropout}, \citet{deepensembles}, and \citet{han2022card}, is adopted. The dataset information in terms of the sample size and number of features is provided in Table~\ref{tab:reg_uci_dim}. For both Kin8nm and Naval dataset, the response variable is scaled by $100$.

The standard $90\%/10\%$ train-test splits in \citet{pbp} ($20$ folds for all datasets except $5$ for Protein and $1$ for Year) is applied, and metrics are summarized by their mean and standard deviation (except Year) across all splits. We compare the performance of DBT to all aforementioned BNN frameworks: PBP, MC Dropout, and Deep Ensembles, plus another deep generative model for learning conditional distributions, GCDS \citep{jointmatching}, as well as CARD. Following the same paradigm of BNN model assessment, we evaluate the accuracy and predictive uncertainty estimation of DBT, CARD and GCDS by reporting RMSE and NLL. Furthermore, we compute QICE for all methods to evaluate distribution matching.

\begin{table}[ht]
\vspace{3mm}
\caption{\label{tab:reg_uci_dim}Dataset size ($N$ observations, $P$ features) of UCI regression tasks.}\vspace{-2.5mm}
\begin{center}
\resizebox{\textwidth}{!}{
\begin{tabular}{@{}c|cccccccccc@{}}
\toprule[1.5pt]
Dataset  & Boston      & Concrete     & Energy    & Kin8nm       & Naval          & Power        & Protein       & Wine          & Yacht     & Year            \\ \midrule
$(N, P)$ & $(506, 13)$ & $(1030,8)$ & $(768,8)$ & $(8192,8)$ & $(11,934,16)$ & $(9568,4)$ & $(45,730,9)$ & $(1599,11)$ & $(308,6)$ & $(515,345,90)$     \\ \bottomrule[1.5pt]
\end{tabular}}
\end{center}
\vspace{5mm}
\end{table}

\subsection{A Closer Look at DBT's Performance in UCI Regression Tasks}
\label{ssec:performance_on_uci_reg}
For the UCI regression tasks, DBT trained on the full dataset performs on par with CARD in most cases, while outperforming other baseline methods. We echo two insightful observations from the seminal paper on gradient boosting by \citet{friedman2001gbm}:

\begin{itemize}[noitemsep, topsep=0pt, leftmargin=*]
    \item ``The performance of any function estimation method depends on the particular problem to which it is applied.''
    \item ``Every method has particular targets for which it is most appropriate and others for which it is not.''
\end{itemize}

We also encourage readers to review Table 1 in \citet{deepensembles}: the metrics are marked in bold by taking into account the error bars. By applying the convention in our work --- where only \textbf{the metrics with the best mean} are marked in bold --- only 2 out of 10 datasets would feature bold metrics for their proposed method (Deep Ensembles) in terms of RMSE, as opposed to the 8 out of 10 reported.

We emphasize that this is the inaugural work on \textbf{Diffusion Boosting}, and our primary goal is to introduce this framework as the first model that 1) is simultaneously a diffusion-based generative model and a boosting algorithm, and 2) can be parameterized by trees to model a conditional distribution, without any assumptions on its distributional form. We have highlighted DBT's advantage over CARD in modeling piecewise-defined functions (\cref{sssec:reg_toy}) and, although it secures slightly fewer Top-2 results than CARD on the UCI datasets, it already outperforms all other baseline methods. These outcomes convincingly demonstrate the potential of our proposed framework. We believe that the results endorse the framework's capabilities, and we look forward to further enhancements in future work.

\subsection{Assessing the Efficacy of An Amortized GBT: One Model for All Timesteps}
\label{ssec:amortized_gbt_ablation_study}
We replaced the noise-predicting network in CARD with an amortized GBT model and ran the experiment on UCI benchmark datasets. Despite extensive tuning, we observed consistently poor performance across all datasets. For example, on the Boston dataset, the best results we obtained were: RMSE $24.76 \pm 1.59$, NLL $6.65 \pm 0.00$, and QICE $16.98 \pm 0.01$, which are significantly worse than those reported in Table~\ref{tab:reg_uci_metric_tables}. For the GBT model, we used $1,000$ trees, $10,000$ noise samples for each instance, $31$ leaves per tree, with timestep $t$ as an additional model input.

We believe this discrepancy can be attributed to several key factors:

\begin{itemize}[noitemsep, topsep=0pt, leftmargin=*]
    \item  Our experiment involved drawing $10,000$ noise samples for each instance, paired with a randomly sampled timestep. Consequently, on average, each instance was only paired with $10$ noise samples per timestep --- far fewer than our standard hyperparameter setting of $100$ noise samples per tree. Increasing the number of noise samples is impractical, due to the substantial memory requirements incurred by duplicating the dataset multiple times. Notably, the UCI Boston dataset, one of the smallest datasets as shown in \cref{tab:reg_uci_dim}, contains fewer than $500$ training samples.
    \item A tree model only has as many distinct outputs as the number of leaves, thus struggles to accommodate the diversity of outputs required across all $1,000$ diffusion timesteps to make good predictions.
\end{itemize}

These challenges highlight the limitations of using a single amortized GBT model under our experimental conditions, and substantiate our choice to employ a different tree at each timestep in our study.

\subsection{Runtime Performance Analysis}
\label{ssec:runtime_analysis}
We implemented DBT using the official PyTorch repository of CARD to directly leverage their model evaluation framework. (Therefore in our experiments, when the dataset does not contain missing values, the conditional mean estimator $f_{\phi}(\vx)$ is parameterized by deep neural networks out of convenience.)

During training, the time required to train each tree is relatively short. For instance, on our largest dataset, UCI Year, which has a training set dimensionality of $(46,371,500, 90)$, it takes approximately 100 seconds to train each tree using an AMD EPYC 7513 CPU. In contrast, on a smaller dataset like UCI Boston with a dimensionality of $(45,500, 13)$, each tree takes about 0.03 seconds to train. However, constructing the training set for each tree consumes more time; for the UCI Year dataset, this process takes about 2.5 minutes.

During inference, the procedure is inherently sequential as each tree's output is required to construct the Gaussian mean for the sampling of $\vy$ in the next diffusion timestep: see Eq.\,(\ref{eq:card_post}) and (\ref{eq:card_post_mean}). This sequential nature prevents parallelization during inference, setting it apart from other contemporary gradient boosting algorithms.

Given our focus on methodology development in this work, we intend to explore system optimizations, including runtime performance, in our future research.

\section{Limitations}
\label{sec:limitations}
As discussed in \cref{sssec:reg_uci} and \cref{ssec:performance_on_uci_reg}, our method aims to tackle tabular data, which is inherently heterogeneous. Consequently, it is challenging to identify a set of tabular datasets that adequately represent the diversity of such data. In this work, we follow the practice of \citet{han2022card} by testing our method on $10$ UCI regression datasets (\cref{tab:reg_uci_dim}). These datasets are standard benchmarks in many works, including those focused on predictive uncertainty, facilitating easier benchmarking with existing methods. However, we note that half of these datasets are relatively small, which might not fairly reflect model performance in terms of quantile-based evaluation metrics that measure distribution matching, such as QICE.

Additionally, as mentioned at the beginning of \cref{sec:experiments}, our model currently requires batch learning instead of mini-batch learning. The need for multiple noise samples per instance necessitates duplicating the entire training dataset multiple times, resulting in significant memory consumption compared to CARD or other neural network-based methods. We emphasize that this limitation arises from the choice of package (LightGBM) rather than our method itself. We have identified a potential solution using the data iterator functionality in XGBoost and plan to address this limitation in future work.

Finally, as discussed in \cref{ssec:runtime_analysis}, unlike other contemporary gradient boosting libraries, our model’s prediction computation cannot be parallelized due to the sequential nature of the reverse process sampling. This limitation constrains the evaluation speed during inference.

\section{Broader Impacts}
\label{sec:impacts}
We discussed in \cref{sssec:ota_fraud_detection} the deployment of DBT as a model for learning to defer, which could potentially have positive societal impacts by enhancing decision-making through human-AI collaboration for anomaly detection. While the proposed framework offers promising solutions for tackling supervised learning problems, several potential negative societal impacts need to be considered. Privacy issues may arise if the response variable samples generated by the model inadvertently leak sensitive information or enable the re-identification of individuals in anonymized datasets. Additionally, although our proposed framework in \cref{ssec:clf} aims to promote human-in-the-loop business scenarios, the improved decision-making capabilities might still lead to increased automation in industry settings, potentially displacing workers in roles involving routine decision-making tasks. Finally, the environmental impact of training computationally expensive diffusion models should not be overlooked, as it contributes to increased energy consumption and a larger carbon footprint.

\end{document}